\documentclass[letterpaper]{article} 
\usepackage[]{aaai2026}  
\usepackage{times}  
\usepackage{helvet}  
\usepackage{courier}  
\usepackage[hyphens]{url}  
\usepackage{graphicx} 
\usepackage{natbib}  
\usepackage{caption} 
\usepackage{algorithm}
\usepackage{algorithmic}
\usepackage{amssymb}
\usepackage{amsmath}
\newtheorem{definition}{Definition}
\newtheorem{lemma}{Lemma}
\usepackage{enumitem}
\usepackage{tabularx}
\usepackage{booktabs}
\usepackage{tabulary}
\usepackage{multirow}
\usepackage{makecell}
\usepackage{subcaption}
\usepackage{longtable}
\usepackage{hyperref}
\usepackage{xcolor} 
\usepackage{makecell}
 
\title{AMaPO: Adaptive Margin-attached Preference Optimization for Language Model Alignment}
\author {
    Ruibo Deng\textsuperscript{\rm 1,2},
    Duanyu Feng\textsuperscript{\rm 1,2},
    Wenqiang Lei\textsuperscript{\rm 1,2}\thanks{Corresponding author}
}
\affiliations{
    \textsuperscript{\rm 1}Sichuan University, Chengdu, China\\
    \textsuperscript{\rm 2}Engineering Research Center of Machine Learning and Industry Intelligence, Ministry of Education, China \\
    dengruibo@stu.scu.edu.cn, fengduanyuscu@stu.scu.edu.cn, wenqianglei@scu.edu.cn
}

\begin{document}

\maketitle
\frenchspacing

\begin{abstract}
Offline preference optimization offers a simpler and more stable alternative to RLHF for aligning language models. However, their effectiveness is critically dependent on ranking accuracy, a metric where further gains are highly impactful. This limitation arises from a fundamental problem that we identify and formalize as the Overfitting-Underfitting Dilemma: current margin designs cause models to apply excessive, wasteful gradients to correctly ranked samples (overfitting) while providing insufficient corrective signals for misranked ones (underfitting). To resolve this dilemma, we propose \textbf{Adaptive Margin-attached Preference Optimization (AMaPO)}, a simple yet principled algorithm. AMaPO employs an instance-wise adaptive margin, refined by Z-normalization and exponential scaling, which dynamically reallocates learning effort by amplifying gradients for misranked samples and suppressing them for correct ones. Extensive experiments on widely used benchmarks demonstrate that AMaPO not only achieves better ranking accuracy and superior downstream alignment performance, but targeted analysis also confirms that it successfully mitigates the core overfitting and underfitting issues. \footnote{Our code: \url{https://github.com/Shiroha-Offical/AMaPO}}
\end{abstract}


\section{Introduction}
Aligning Large Language Models (LLMs) with human preferences, ensuring they are helpful, honest, and harmless, also known as preference learning, is crucial for LLM research \cite{bai2022training, ouyang2022training}. A standard approach for this task is Reinforcement Learning from Human Feedback (RLHF), a powerful online algorithm that first trains a reward model on human preferences and then uses reinforcement learning to optimize the LLMs \cite{ouyang2022training, schulman2017proximalpolicyoptimizationalgorithms}. Despite its demonstrated success, RLHF is notoriously complex and prone to training instabilities, presenting significant practical challenges \cite{engstrom2020implementation}. To circumvent these issues, a new family of offline alignment algorithms has emerged, led by methods like DPO \cite{rafailov2023direct}, IPO \cite{azar2024general}, KTO \cite{ethayarajh2024ktomodelalignmentprospect} and FocalPO \cite{liu2025focalpo}. These approaches offer superior simplicity and stability by directly optimizing the LLMs to favor preferred responses over dispreferred ones, thereby bypassing the need for an explicit reward model and the complexities of RL. At its core, the efficacy of this entire paradigm is the ranking accuracy of an implicit reward model, which dictates how well the optimization process can distinguish between desirable and undesirable outputs.

Therefore, many subsequent DPO-style methods have been used to improve ranking accuracy, either explicitly prioritizing ranking \cite{chen2024preference} or indirectly fostering this capability through other modifications. A primary direction of these methods is to manipulate the implicit reward function, often by incorporating a reward margin to better separate preferred and dispreferred responses. Innovations in this direction include introducing fixed or dynamic margins \cite{meng2024simpo, wu2024alpha} or updating the reference model to create more favorable optimization landscapes \cite{gorbatovski2024learn}. Although these methods report improved performance, their contributions are typically validated empirically, without a unified theoretical perspective to explain how these methods dynamically affect ranking accuracy during training and to compare with each others. Separately, emerging work has analyzed the gradient dynamics of preference learning algorithms \cite{feng2024towards, yan20243d, yuan2024common, ma2025gradient}. However, these studies often fail to explicitly connect their findings on gradient properties, such as magnitude or conflict, with the resulting changes in ranking accuracy. Therefore, we argue that a crucial analytical bridge is missing: a framework that not only examines the gradient dynamics of these advanced DPO variants but also directly links them to the dynamic evolution of ranking accuracy.

To bridge this analytical gap, we introduce a unified framework that analyzes DPO-style methods through the lens of the reward margin. First, we reformulate existing algorithms, including DPO and its variants, into a generalized objective function based on different margin designs. This unification establishes a direct, formal connection between the algorithmic structure of each method and its capacity to optimize ranking accuracy. Within this framework, we then employ gradient dynamics analysis to investigate how these distinct margin formulations shape the learning trajectory of ranking accuracy. Our analysis reveals a fundamental flaw in the standard DPO mechanism: for samples that are already correctly ranked, its margin design leads to \textit{overfitting} by assigning excessively large gradients, wasting capacity on these easy samples. Conversely, for incorrectly ranked samples where learning is most critical, the margin often results in \textit{underfitting} by providing insufficient gradients, thereby hindering the correction of errors. We also find that while some of subsequent DPO-style methods attempt to address this, they fail to fully resolve this core tension.

Motivated by our analysis, we introduce AMaPO, an algorithm designed to resolve this overfitting-underfitting dilemma. AMaPO's core strategy is an instance-wise adaptive margin, refined by Z-normalization and exponential scaling. By dynamically assigning large corrective margins to underfit samples and a zero margin to correctly ranked ones, it effectively reallocates learning effort to where it is most needed. Our extensive experiments on diverse benchmarks demonstrate that AMaPO not only achieves state-of-the-art ranking accuracy and superior downstream alignment performance, but targeted analysis also confirms that it successfully mitigates the core overfitting and underfitting issues, validating our approach.

The main contributions of this paper are: (i)  We introduce a unified margin-based framework to analyze gradient dynamics of DPO-style algorithms, which reveals an overfitting-underfitting dilemma: existing methods expend excessive gradient on already-learned preferences (overfitting) while providing insufficient signal to correct misranked ones (underfitting). (ii) We propose Adaptive Margin-attached Preference Optimization (AMaPO), a principled algorithm designed to resolve this dilemma. AMaPO employs an instance-wise adaptive margin and refined by Z-normalization and exponential scaling, that reallocates learning effort by amplifying gradients for misranked samples and suppressing them for correctly ranked ones. (iii) Extensive experiments demonstrate AMaPO achieves state-of-the-art ranking accuracy, solving the overfitting-underfitting dilemma and leading to superior downstream alignment performance like instruction-following and complex reasoning.
\section{Related works}
\label{sec:related_works}
\paragraph{Preference Learning Algorithms}
The field of preference learning for LLM alignment is dominated by two primary approaches: online and offline. Online methods, exemplified by Reinforcement Learning from Human Feedback (RLHF) \cite{bai2022training, ouyang2022training}, employ a complex two-stage process of reward modeling and RL-based optimization. Offline methods, initiated by Direct Preference Optimization (DPO) \cite{rafailov2023direct}, directly optimize the LLM on preference data, obviating practical challenges such as the need for a separate reward model and training instability. Optimized DPO variants generally fall into two categories: reformulating the core loss function \cite{azar2024general, ethayarajh2024ktomodelalignmentprospect}, or incorporating an explicit reward margin to separate preference pairs robustly \cite{boser1992training, turner2012bradley}. This margin-based approach has been realized through employing fixed or instance-dependent margins to improve learning stability and handle potential data noise \cite{meng2024simpo, kim2024margin, amini2024direct}. Others achieve a form of dynamic regularization by progressively updating the reference policy itself, which effectively creates an adaptive margin throughout the training process \cite{gorbatovski2024learn, wu2024alpha}.
Despite these diverse, the central mechanism common to all these DPO-style algorithms is the implicit reward function. This makes their ability to learn to classify preferences correctly, known as the ranking accuracy \cite{chen2024preference}, the paramount determinant of their success. However, the field currently lacks a unified framework to analyze how these different methods impact the evolution of ranking accuracy, leaving a critical gap in our understanding.

\paragraph{Preference Learning Theory}
Concurrent with the development of preference learning algorithms, a body of theoretical work has emerged to analyze their underlying mechanisms. These inquiries generally follow two primary perspectives: gradient dynamics and divergence analysis. From the gradient perspective, for example, \cite{feng2024towards} reveals the limited learning capacity of DPO through field theory and \cite{yuan2024common} identify issues such as gradient entanglement in margin-based methods. From divergence perspective, it examines how DPO optimizes the model distribution $\pi_\theta$ towards the distribution of chosen response. While DPO implicitly optimizes a forward KL-divergence, recent work identifies promoting mode-seeking behavior by optimizing a reverse KL-divergence or Total Variation distance as a crucial property for preference alignment\cite{tajwar2024preference, xiao2024cal, xiao2025simper}. While these analyses are foundational, they typically examine each algorithm as a distinct entity. In this paper, we also want to build upon the tradition of gradient analysis but introduce a novel, unifying perspective. Based on the unification, we further explicitly connect the specific gradient properties of each method to their dynamic impact on ranking accuracy.
\section{Preliminaries}
This section establishes the technical foundations for our analysis. First, we detail Direct Preference Optimization (DPO) \cite{rafailov2023direct}, the seminal algorithm for offline preference learning that serves as the basis for the methods we investigate. Second, we formalize the concept of the reward margin, which is used to construct a unified representation of various DPO-style algorithms. Finally, we define ranking accuracy, the primary metric used throughout our work to evaluate the efficacy of these alignment methods.

\subsection{Direct Preference Optimization}

\paragraph{Preference Data and the Bradley-Terry Model.}
Offline preference optimization relies on a static dataset of preferences, $\mathcal{D} = \{ (x^{(i)}, y_w^{(i)}, y_l^{(i)}) \}_{i=1}^N$. Each entry consists of a prompt $x^{(i)}$ and a pair of responses, $(y_w^{(i)}, y_l^{(i)})$, where $y_w$ is preferred over $y_l$, denoted as $y_w \succ y_l$. This preference is typically provided by a human annotator or a powerful reward model. A standard assumption in the field is that these preferences are drawn from a distribution that follows the Bradley-Terry (BT) model \cite{bradley1952rank}, which models the probability of preferring $y_w$ over $y_l$ as a logistic function of the difference in their latent reward values: $p(y_w \succ y_l | x) = \sigma(r(x, y_w) - r(x, y_l))$.

\paragraph{The Direct Preference Optimization.}
Instead of explicitly training a reward model $r(x,y)$, Direct Preference Optimization (DPO) \cite{rafailov2023direct} derives an analytical mapping from the optimal policy to the reward function. This allows DPO to use the log-likelihood of the policy to implicitly represent the reward function via a closed-form expression with the optimal policy:
\begin{align}
\mathcal{L}_{\mathrm{DPO}}(\pi_{\theta};\mathcal{D}) 
= &-\mathbb{E}_{(x, y_w, y_l) \sim \mathcal{D}} \big[ \log \sigma \big( \beta\log \tfrac{\pi_\theta(y_w | x)}{\pi_{\mathrm{ref}}(y_w | x)} \nonumber \\
&- \beta\log \tfrac{\pi_\theta(y_l | x)}{\pi_{\mathrm{ref}}(y_l | x)} \big) \big],
\end{align}
where $\pi_{\mathrm{ref}}$ is a fixed reference model, and $\beta$ is a temperature parameter that scales the reward.

\subsection{A Unified Margin-based Framework}
\label{sec:margin_framework}

Recent theoretical works have sought to unify DPO-style algorithms under a single mathematical representation \cite{yuan2024common, liu2024understanding}. A common approach is to formulate a general loss function that captures variations in reward shaping and regularization. However, with the increasing prominence of methods that explicitly engineer a reward margin \cite{zhao2024rainbowpo, wu2024alpha}, these existing frameworks can represent such methods only indirectly, obscuring the margin's central role.

To create a more direct analytical tool that reflects this research trend, we extend the prior work \cite{yuan2024common} by proposing a refined unified objective with the margin, $\gamma$:
\begin{align}
    \mathcal{L}_{\mathrm{unified}}(\theta)=&-(m(h_w(\log \pi_w)-h_l(\log \pi_l) - \gamma) \nonumber \\ 
    &+ \Lambda\left(\log \pi_w\right)).
\label{unified_framework}
\end{align}
Here, $h_w(x)$ and $h_l(x)$ represent transformation or adjustment functions applied to the log-probabilities of the preferred and dispreferred responses, respectively. 
$\pi_w$ and $\pi_l$ represent $\pi_\theta\left(y_w | x\right)$ and $\pi_\theta\left(y_l | x\right)$ for simplicity, respectively.
The term $m(x)$ is a scoring function that computes the difference between these transformed values with the margin $\gamma$. Finally, $\Lambda\left(\log \pi_w\right)$ is an optional auxiliary term that some models incorporate to encourage additional learning specifically on the preferred response.
This formulation allows us to directly compare algorithms based on their margin design. For instance:

\textbf{DPO}: The original DPO algorithm fits this framework with $h_w(x)=h_l(x)=\beta x$, $m(x)=\log \sigma(x)$ and a fixed margin of $\gamma = \beta(\log \pi_{\mathrm{ref}}(y_w|x) - \log \pi_{\mathrm{ref}}(y_l|x))$.

\textbf{SimPO}: Simple Preference Optimization (SimPO) \cite{meng2024simpo} is a highly effective DPO variant, shown as
\begin{align} 
\mathcal{L}_{\mathrm{SimPO}}(\pi_{\theta};\mathcal{D})
= &-\mathbb{E}_{\left(x, y_w, y_l\right) \sim \mathcal{D}} \big[ \log \sigma \big(\tfrac{\beta}{|y_w|}\log\pi_\theta\left(y_w | x\right) \nonumber \\ 
&- \tfrac{\beta}{|y_l|} \log \pi_\theta\left(y_l | x\right) - C \big)\big].  
\label{SimPO_loss}
\end{align}
It fits our framework by defining $h_w(x)=\frac{\beta}{|y_w|}x $, $h_l(x)=\frac{\beta}{|y_l|} x$, $m(x)=\log \sigma(x)$ and a tunable margin $\gamma = C > 0$.

This margin-focused unification serves as the cornerstone of our gradient analysis. \footnote{For completeness, we formulate other common DPO-style methods within this framework in Appendix~\ref{sec:baseline_methods}.}

\subsection{Ranking Accuracy}

While the ultimate goal of alignment is to produce generations that humans prefer, which is often evaluated by win-rate against a baseline model \cite{zheng2023secrets}, this metric is costly and ill-suited for tracking training dynamics. Therefore, a more direct and practical metric is crucial for analyzing the learning process. Ranking accuracy \cite{liu2024rm, chen2024preference} directly measures whether the policy (training) model $\pi_\theta$ itself assigns a higher likelihood to the preferred response over the dispreferred one for a given prompt. Formally, it is defined as:
\begin{equation}
    \mathcal{R}(\pi_\theta; \mathcal{D})=\frac{1}{|\mathcal{D}|} \sum_{\left(x, y_w, y_l\right) \in \mathcal{D}} \mathbb{I} \left[\pi_\theta(y_w|x)>\pi_\theta(y_l|x)\right],
    \label{eq:ranking_accuracy}
\end{equation}
where $\mathbb{I}[\cdot]$ is the indicator function. A higher ranking accuracy indicates that the policy's preference ordering aligns with the ground-truth data, and it serves as a predictor of the downstream performance within a certain KL divergence range \cite{chen2024preference}. Furthermore, for our theoretical analysis, we utilize the instance-wise ranking accuracy derived from our unified framework (Eq.~\ref{unified_framework}), defined as $h_w(\log \pi_w) - h_l(\log \pi_l)$. For the vast majority of DPO-style methods where the transformations $h_w$ and $h_l$ are linear or monotonic, the sign of this score is equivalent to the standard ranking accuracy (i.e., $h_w(\log \pi_w) - h_l(\log \pi_l) > 0 \iff \pi_\theta(y_w|x) > \pi_\theta(y_l|x)$). Throughout this paper, we use this ranking accuracy as the primary lens for analysis.
\section{Methods}

\setlength{\tabcolsep}{3mm}
\begin{table*}[!t]
\centering
  \small
    \caption{All four possible cases for training dynamics of the unified framework in Eq. \ref{unified_framework}, where $\uparrow$ and $\downarrow$ indicate increase and decrease, $r_{\pi_\theta}(x,y_w,y_l)=h_w(\log \pi_w) - h_l(\log \pi_l)$. The \textbf{correctly ranked} means $r_{\pi_\theta}(x,y_w,y_l)>0$, the \textbf{incorrectly ranked} means $r_{\pi_\theta}(x,y_w,y_l)\leq 0 $. \textbf{Case 1} and \textbf{Case 4} are the ideal behavior which properly control the learning rate.}
  \label{case_analysis}
  \vspace{-2pt}
  \centering
    \renewcommand{\arraystretch}{0.8}
    \begin{tabular}{cccl}
        \toprule
        \textbf{Case} & $r_{\pi_\theta}(x,y_w,y_l)$, $\gamma$ & $|d_{\boldsymbol\theta}|$ & \textbf{Influence} \\ 
        \midrule
        1    & $r_{\pi_\theta}(x,y_w,y_l)\geq\gamma$ & $\downarrow$ & mild update on \textbf{correctly ranked} samples, \textbf{ideal}.\\ 
        \midrule
        2    & $r_{\pi_\theta}(x,y_w,y_l)<\gamma$ & $\uparrow$ & aggressive update on \textbf{correctly ranked} samples, \textbf{overfitting}.\\ 
        \midrule
        3    & $r_{\pi_\theta}(x,y_w,y_l)\geq\gamma$ & $\downarrow$ & mild update on \textbf{incorrectly ranked} samples, \textbf{underfitting}.\\ 
        \midrule
        4    & $r_{\pi_\theta}(x,y_w,y_l)<\gamma$& $\uparrow$ & aggressive update on \textbf{incorrectly ranked} samples, \textbf{ideal}.\\ 
        \bottomrule
    \end{tabular}
    \vspace{-10pt}
\end{table*}

In this section, we leverage the unified margin-based framework established in the previous section to conduct a detailed theoretical analysis and construct our related method. Our primary goal is to understand how the margin design in DPO and its variants affects the learning dynamics of ranking accuracy. Through this gradient-level investigation, we diagnose the key limitations of existing approaches and, from these findings, distill a set of desiderata for a more effective margin function. Building on these principles, we then introduce \textbf{Adaptive Margin-attached Preference Optimization (AMaPO)}, a novel preference optimization algorithm that assigns an adaptive margin specifically designed to manipulate gradient magnitude, thereby improving both ranking accuracy and overall alignment performance.

\subsection{Gradient Dynamics and the Ideal Margin}
\label{sec:gradient_analysis}

Based on the unified objective in Eq.~\ref{unified_framework}, we can derive a general gradient formulation for DPO-style algorithms. The gradient with respect to the model parameters $\boldsymbol{\theta}$ is:
\begin{align}
    \label{eq:gradient_full}
    \nabla_{\boldsymbol{\theta}} \mathcal{L}_{\mathrm{unified}}&\left(\theta\right) = -m^{\prime}(h_w(\log \pi_w) - h_l(\log \pi_l) - \gamma) \nonumber \\
    &(h_w^{\prime}(\log \pi_w) - h_l^{\prime}(\log \pi_l)) - \Lambda^{\prime}\left(\log \pi_w\right).
\end{align}
Specifically, we define $d_{\boldsymbol\theta}=m^{\prime}(h_w(\log \pi_w)-h_l(\log \pi_l)-\gamma)$, which comprises log-probabilities of responses ($h_w(\log \pi_w)$, $h_l(\log \pi_l)$) and margin $\gamma$ and significantly affects the gradient magnitude\footnote{For simplicity, we omit the contribution of $\Lambda^{\prime}(\log \pi_w)$ to the gradient magnitude, as $\Lambda^{\prime}(\log \pi_w)$ is typically a constant and not related to the $\gamma$.}. 

This insight reveals a crucial mechanism: for a given state of the policy $\pi_\theta$, the \textbf{margin $\gamma$ is the primary lever to control the learning rate}. Since the gradient magnitude $d_{\boldsymbol\theta}$ is governed by 
\begin{equation}
    d_{\boldsymbol\theta}=m^{\prime}(h_w(\log \pi_w)-h_l(\log \pi_l)-\gamma)
    \label{gradient_magnitude}
\end{equation}
where $m^{\prime}$ is usually a monotonically decreasing function and $m^{\prime}(x)>0$ (as $m(x)=\log \sigma(x)$), the choice of $\gamma$ directly determines whether the learning signal is amplified or suppressed. A poorly designed margin leads directly to what we term the Overfitting-Underfitting Dilemma.

\begin{definition}[The Overfitting-Underfitting Dilemma]
\label{def:dilemma}
An algorithm is susceptible to this dilemma when its margin $\gamma$ causes inefficient optimization:
\begin{itemize}[leftmargin=*, labelindent=0pt]
    \item \textbf{Overfitting:} A large margin produces wasteful gradients for correctly ranked samples, forcing the model to over-optimize on these samples.
    \item \textbf{Underfitting:} A small margin yields small gradients for incorrectly ranked samples, stifling learning where it is most needed.
\end{itemize}
\end{definition}

Table~\ref{case_analysis} provides a concrete illustration of these cases. Resolving this dilemma requires a margin that adheres to a key principle: it must be \textbf{dynamically adaptive} to the policy's instance-wise ranking accuracy. This motivates the concept of an Oracle Ranking Margin.

\begin{definition}[Oracle Ranking Margin]
\label{def:oracle_margin}
The \textbf{Oracle Ranking Margin}, $\gamma^*$, is an instance-specific, non-negative threshold that acts as a dynamic learning target \footnote{We provide further theoretical analysis in Appendix~\ref{app:analysis_oracle_margin}}, distinguishing between incorrectly ranked samples that require a strong learning signal $r_{\pi_\theta}(x,y_w,y_l) < \gamma^*$, and correctly ranked ones whose gradients should be suppressed to prevent overfitting $r_{\pi_\theta}(x,y_w,y_l) > \gamma^*$. The instance-wise ranking accuracy $r_{\pi_\theta}(x,y_w,y_l)$ is defined as $h_w(\log \pi_w)-h_l(\log \pi_l)$.
\end{definition}

With these principles established, to thoroughly investigate whether existing DPO-style algorithms satisfy the above demands, our analysis focuses on two of the most celebrated objectives, including DPO and SimPO \footnote{The analysis of other methods is in Appendix \ref{Gradient Analysis}.}.

\paragraph{Analysis of DPO.} The margin, $\gamma = \beta\log \pi_{\mathrm{ref}}(y_w|x) - \beta \log \pi_{\mathrm{ref}}(y_l|x)$ derived from the reference model, can neither dynamically adapt to instance-wise ranking correctness nor remaining positive. This irrelevance suggests that DPO can reproduce all cases above, which might result in overfitting and underfitting, as illustrated by case 2 and case 3.

\paragraph{Analysis of SimPO.} Although $\gamma = C>0$ is guaranteed to remain positive, it overlooks the variability inherent in instance-wise ranking correctness. This rigidity suggests that SimPO can reproduce case 2 and case 4, which could lead to overfitting as shown in case 2.

Our analysis reveals that margin designs of both DPO and SimPO fail to proactively adapt based on instance-wise ranking correctness, leading to overfitting and underfitting in preference learning and, consequently, suboptimal ranking accuracy and alignment performance.

\subsection{AMaPO: Adaptive Margin-attached Preference Optimization}

To overcome the aforementioned limitations, we introduce \textbf{Adaptive Margin-attached Preference Optimization (AMaPO)}.
Our approach is designed to directly resolve the Overfitting-Underfitting Dilemma by dynamically adjusting the gradient magnitude for each sample. We build upon the robust SimPO \cite{meng2024simpo} framework, but our key innovation lies in replacing its static margin with a principled, instance-wise adaptive margin, $\gamma(x,y_w,y_l)$. The general objective is:
\begin{align}
\mathcal{L}_{\mathrm{AMaPO}}(\pi_\theta;\mathcal{D})=&-\mathbb{E}_{\left(\mathbf{x}, \mathbf{y}_w, \mathbf{y}_l\right) \sim \mathcal{D}} \big[\log \sigma\big(
r_{\pi_\theta}(x,y_w,y_l) \nonumber \\ 
&- \gamma(x,y_w,y_l)\big)\big],
\label{AMaPO_unified_framework}
\end{align}
where $r_{\pi_\theta}(x,y_w,y_l) = \frac{\beta}{|y_w|}\log \pi_\theta(y_w|x) - \frac{\beta}{|y_l|}\log \pi_\theta(y_l|x)$ is the implicit ranking accuracy, and $m(x)=\log \sigma(x)$ as  Eq.~\ref{unified_framework}. The following subsections detail the design of our adaptive margin, $\gamma(x,y_w,y_l)$.

\paragraph{Formulating the Ideal Adaptive Margin.}
Our theoretical analysis concluded that an ideal margin must be \textit{dynamically adaptive}. To satisfy this, from definition \ref{def:oracle_margin}, the margin should reflect the gap between the policy's performance and the ideal target set by the Oracle Ranking Margin, $\gamma^*$. We therefore formulate the ideal adaptive margin as:
\begin{equation}
    \gamma^*(x,y_w,y_l) = \mathbb{I}\left[(\gamma^* - r_{\pi_\theta}(x,y_w,y_l))>0\right]\cdot \gamma^*.
    \label{ideal_margin}
\end{equation}
This formulation is inherently dynamic and directly addresses the dilemma. For incorrectly ranked samples where $r_{\pi_\theta} \leq 0 < \gamma^*$, the margin is positive and large, amplifying the corrective gradient to mitigate underfitting. For correctly ranked samples where $r_{\pi_\theta} > \gamma^* \geq 0$, the margin becomes zero, suppressing the gradient and preventing overfitting.

\paragraph{Estimating the Oracle Margin in Practice.}
Since the true Oracle Ranking Margin $\gamma^*$ is inaccessible, we must estimate it. Motivated by empirical findings that the implicit reward distributions of aligned models are often right-skewed \cite{qin2024towards}, we propose using the mean implicit margin $\mu_r$ within the current training batch $B$ as a robust, annotation-free proxy for $\gamma^*$. To make the ideal margin from Eq.~\ref{ideal_margin} tractable by relaxing the non-negativity constraint (i.e., the $\max(\cdot, 0)$ function), while also stabilizing the estimation and scaling it appropriately, we apply Z-score normalization \cite{patro2015normalization}. This yields:
\begin{equation}
    \gamma(x,y_w,y_l) = \max\big(\tfrac{\mu_r - r_{\pi_\theta}(x,y_w,y_l)}{\sigma_r} \cdot \mu_r, 0\big),
    \label{AMaPO_margin}
\end{equation}
where $\mu_r$ and $\sigma_r$ are the mean and standard deviation of $r_{\pi_\theta}(x,y_w,y_l)$ computed within batch $B$. The term $\frac{\mu_r - r_{\pi_\theta}}{\sigma_r}$ calculates the normalized "difficulty" of the sample, which is then scaled by the estimated oracle target $\mu_r$ itself.

\paragraph{Margin Scaling for Quality Representation.}
The margin from Eq.~\ref{AMaPO_margin} effectively captures the relative difficulty of a sample. However, empirical results find that the log probability value might not truthfully reflect the quality of generated sequence \cite{holtzman2021surface}. To better represent the quality gap between responses and speed up the training of the hard incorrectly ranked samples, we introduce a scaling function. Inspired by the strong correlation between perplexity (PPL) and generation quality \cite{marion2023less, gonen2023demystifying}, we use an exponential scaling function. The scaled margin is \footnote{This elegantly transforms our additive, log-space margin into the geometric mean of the PPL ratio within the batch. The derivation is in Appendix \ref{Derivations for Margin Scaling}.}:
\begin{equation}
    h_\gamma(\gamma) = 
    \begin{cases}
        0 & \text{if } \gamma = 0, \\
        \beta \cdot e^\gamma & \text{if } \gamma > 0.
    \end{cases}
    \label{margin_scaling}
\end{equation}

\paragraph{Final Objective.}
Finally, to ensure that our adaptive margin serves as a fixed target for each sample within an optimization step, we apply a stop-gradient operation ($\mathrm{sg}[\cdot]$). This prevents the gradient from flowing back through the margin calculation. Incorporating all components, the final AMaPO objective is:
\begin{align}
    \mathcal{L}_{\mathrm{AMaPO}}(\pi_\theta;\mathcal{D})=&-\mathbb{E}_{\left(\mathbf{x}, \mathbf{y}_w, \mathbf{y}_l\right) \sim \mathcal{D}} \big[\log \sigma\big(
    r_{\pi_\theta}(x,y_w,y_l) \nonumber \\ 
    &-h_\gamma(\mathrm{sg}[\gamma(x,y_w,y_l)])\big)\big].
    \label{AMaPO_final_objective}
\end{align}
In summary, by employing an adaptive margin derived from the model's current ranking correctness, AMaPO effectively manipulates the gradient to focus on learning from its errors. This design directly resolves the overfitting and underfitting issues inherent in prior methods, leading to improved ranking accuracy and superior alignment performance.
\section{Experiments}
This section presents a series of experiments designed to rigorously evaluate AMaPO from multiple perspectives. Our primary goal is to establish its overall superiority against current methods, measured by both ranking accuracy and performance on downstream tasks. Furthermore, we provide a targeted analysis to confirm that AMaPO directly addresses the core overfitting and underfitting issues identified in our framework. Finally, detailed ablation studies to validate the importance of each components.

\subsection{Experimental Setup}
\label{sec:Experimental_setup}
\paragraph{Basemodels.}
To assess the robustness and general applicability of our method, we experiment with two widely-used open-source model families: Llama3-8B \cite{llama3modelcard} and Mistral-7B \cite{jiang2024identifying}. For each family, we test two distinct scenarios: (1) a \textbf{Base} model setup, where we first perform SFT before preference alignment, and (2) an \textbf{Instruct} model setup, where we begin directly with the publicly available instruction-tuned model. 

\setlength{\tabcolsep}{1mm}
\begin{table*}[t]
\centering
\small 
  \caption{AlpacaEval2 \cite{alpaca_eval} and MT-Bench \cite{zheng2023judging} results under the four settings. LC and WR denote length-controlled and raw win rate, respectively. The best results are highlighted in bold.}
  \label{main-table}
  \renewcommand{\arraystretch}{0.8}
  \begin{tabular}{lcccccccccccc}
    \toprule
    \multirow{3}{*}{\textbf{Method}} & \multicolumn{3}{c}{\textbf{Llama3-8B-Base}} & \multicolumn{3}{c}{\textbf{Llama3-8B-Instruct}} & \multicolumn{3}{c}{\textbf{Mistral-7B-Base}} & \multicolumn{3}{c}{\textbf{Mistral-7B-Instruct}} \\
    \cmidrule(r){2-4} \cmidrule(r){5-7} \cmidrule(r){8-10} \cmidrule(r){11-13} 
    & \multicolumn{2}{c}{\textbf{AlpacaEval 2}} & \textbf{MT-Bench} & \multicolumn{2}{c}{\textbf{AlpacaEval 2}} &\textbf{MT-Bench} & \multicolumn{2}{c}{\textbf{AlpacaEval 2}} & \textbf{MT-Bench} & \multicolumn{2}{c}{\textbf{AlpacaEval 2}} & \textbf{MT-Bench}\\
    \cmidrule(r){2-3} \cmidrule(r){4-4} \cmidrule(r){5-6} \cmidrule(r){7-7} \cmidrule(r){8-9} \cmidrule(r){10-10} \cmidrule(r){11-12} \cmidrule(r){13-13}  
    & \textbf{\scriptsize LC} \scriptsize(\%) & \textbf{\scriptsize WR} \scriptsize (\%) & \textbf{\scriptsize GPT-4 Turbo} & \textbf{\scriptsize LC} \scriptsize (\%) & \textbf{\scriptsize WR} \scriptsize (\%) & \textbf{\scriptsize GPT-4 Turbo} &\textbf{\scriptsize LC} \scriptsize(\%) & \textbf{\scriptsize WR} \scriptsize (\%) & \textbf{\scriptsize GPT-4 Turbo} & \textbf{\scriptsize LC} \scriptsize (\%) & \textbf{\scriptsize WR} \scriptsize (\%) & \textbf{\scriptsize GPT-4 Turbo}\\
    \midrule
    SFT & 6.2 & 4.6 & 3.3 & 26.0 & 25.3 & 6.9 & 8.4 & 6.2 & 4.8 & 17.1 & 14.7 & 6.2 \\
    \midrule
    DPO & 18.2 & 15.5 & 6.5 & 40.3 & 37.9 & 7.0 & 15.1 & 12.5 & 5.9 & 26.8 & 24.9 & 6.3\\
    SLiC & 12.3 & 13.7 & 6.3 & 26.9 & 27.5 & 6.8 & 10.9 & 8.9 & 5.8 & 24.1 & 24.6 & 6.5\\
    IPO & 14.4 & 14.2 & 6.5 & 35.6 & 35.6 & 7.0 & 11.8 & 9.4 & 5.5 & 20.3 & 20.3 & 6.4\\
    KTO & 14.2 & 12.4 & 6.3 & 33.1 & 31.8 & 6.9 & 13.1 & 9.1 & 5.4 & 24.5 & 23.6 & 6.4\\
    CPO & 10.8 & 8.1 & 6.0 & 28.9 & 32.2 & 7.0 & 9.8 & 8.9 & 5.4 & 23.8 & 28.8 & 6.3\\
    SimPO & 22.0 & 20.3 & 6.6 & 44.7 & 40.5 & 7.0 & 21.5 & \textbf{20.8} & 6.0 & 32.1 & 34.8 & 6.6\\
    $\alpha$-DPO & 21.7 & 20.6 & \textbf{6.8} & \textbf{46.6} & 39.6 & \textbf{7.2} & 17.2 & 13.0 & \textbf{6.2} & 34.2 & 33.8 & \textbf{6.7}\\
    \midrule 
    AMaPO & \textbf{26.4} & \textbf{21.4} & 6.4 & 46.1 & \textbf{41.3} & \textbf{7.2} & \textbf{24.3} & 20.6 & \textbf{6.2} & \textbf{34.5} & \textbf{35.1} & \textbf{6.7}\\
    \bottomrule
  \end{tabular}
\end{table*}

\paragraph{Training Datasets.}
Our training pipeline for the \textbf{Base} models follows the well-established recipe \cite{tunstall2023zephyr}: we fine-tune on UltraChat-200k \cite{ding2023enhancing} to create an SFT model, which then is used to for preference optimization on the UltraFeedBack Binarized dataset \cite{cui2023ultrafeedback}. For the \textbf{Instruct} setup, we need to mitigate the distribution shift between the pre-trained instruction model and the preference data \cite{meng2024simpo}. To ensure a controlled and fair comparison, we regenerate the preference pairs by using the Instruct model itself to generate five candidate responses for each prompt from UltraFeedback. We then use a strong reward model, PairRM \cite{jiang2023llm}, to select the best and worst responses, thereby creating a on-distribution preference dataset for the model being trained.

\paragraph{Evaluation benchmarks. } 
\label{sec:evaluation}
Our evaluation is designed to be multi-faceted, assessing performance from ranking accuracy to its practical downstream impact.
(1) \textbf{Ranking Accuracy:} To directly validate our claims about improving preference learning, we measure ranking accuracy on RM-Bench \cite{liu2024rm}, a challenging benchmark designed to test a model's grasp of subtle preference nuances. 
(2) \textbf{Overfitting-underfitting Problem:} To further test our hypothesis about resolving this problem, we also evaluate ranking accuracy across four generalization scenarios on UltraFeedBack following \cite{hong2025robustness}: In-Distribution (ID), Prompt-OOD, Response-OOD, and Mutual-OOD.
(3) \textbf{Downstream Performance:} To confirm that improved ranking translates to better downstream performance, we evaluate on the popular AlpacaEval 2 \cite{alpaca_eval} and MT-Bench \cite{zheng2023judging} benchmarks.

\paragraph{Baselines.}
We compare AMaPO against a comprehensive suite of strong and recent offline preference optimization methods, including DPO \cite{rafailov2023direct}, SLiC \cite{zhao2023slic}, IPO \cite{azar2024general}, KTO \cite{ethayarajh2024ktomodelalignmentprospect}, and SimPO \cite{meng2024simpo}. To ensure a fair and rigorous comparison, we have thoroughly tuned the hyperparameters for each baseline method and report their best achievable performance. \footnote{Detailed experimental setups are provided in Appendix~\ref{sec:Implementation_Details}.}

\setlength{\tabcolsep}{2mm}
\begin{table}[]
    \centering
    \small
    \caption{RM-bench \cite{liu2024rm} results under the three setups, where Easy, Normal and Hard represent the preferred responses have the better, same and worse style compared to less preferred ones. Our AMaPO can achieve good performance on distinguishing responses across various setups.}
  \label{RM-bench-table}
  \renewcommand{\arraystretch}{0.8}
    \begin{tabular}{llcccc}
        \toprule[1pt]
        & \textbf{Method} & \textbf{Avg.}  & Easy & Normal & Hard\\
        \midrule[0.5pt]
        \multirow{4}{*}{\makecell[c]{Mistral-7B \\ Base}}
        & DPO & 55.8 & \textbf{89.8} & 58.7 & 18.7 \\
        & SimPO & 56.5 & 86.3 & 58.7 & 23.5 \\
        & $\alpha$-DPO & \textbf{58.4} & 86.6 & \textbf{62.7} & 25.0 \\
        & AMaPO & 58.1 & 86.4 & 62.5 & \textbf{25.4} \\
         \midrule[0.8pt]
        \multirow{4}{*}{\makecell[c]{Llama3-8B \\ Base}}
        & DPO & 54.6 & \textbf{91.6} & 57.4 & 15.0 \\
        & SimPO & 56.9 & 87.5 & 60.2 & 23.3 \\
        & $\alpha$-DPO & 58.2 & 87.1 & 62.1 & \textbf{25.6} \\
        & AMaPO & \textbf{58.6} & 87.8 & \textbf{62.6} & 25.4 \\
         \midrule[0.8pt]
        \multirow{4}{*}{\makecell[c]{Mistral-7B \\ Instruct}}
        & DPO & 53.2 & 91.3 & 55.5 & 12.9 \\
        & SimPO & 54.9 & 89.0 & 57.8 & \textbf{17.8} \\
        & $\alpha$-DPO & 55.2 & 90.9 & 58.4 & 16.3 \\
        & AMaPO & \textbf{55.5} & \textbf{91.4} & \textbf{59.2} & 15.9 \\
        \midrule
        \multirow{4}{*}{\makecell[c]{Llama3-8B \\ Instruct}}
        & DPO & 53.4 & \textbf{89.6} & 55.4 & 15.1 \\
        & SimPO & 55.7 & 86.1 & 58.7 & 22.4 \\
        & $\alpha$-DPO & 55.0 & 83.4 & 57.6 & 23.8 \\
        & AMaPO & \textbf{56.5} & 85.7 & \textbf{59.8} & \textbf{24.0} \\
    \bottomrule[1pt]
    \end{tabular}
  \vspace{-7pt}
\end{table}


\subsection{Main Results on Benchmarks}
\label{sec:main_results}
\paragraph{Results on Preference Ranking and Downstream Benchmarks.}
Our main results demonstrate that AMaPO achieves superior performance in both preference learning and downstream alignment tasks. 
As shown in Table~\ref{RM-bench-table}, which evaluates the ability to distinguish subtle preference differences, AMaPO consistently outperforms other methods in different scenarios. Specifically, on the Llama3-8B-Base setup, AMaPO improves the ranking accuracy over the strong SimPO baseline by 2.4 and 2.1 points on Normal and Hard cases, respectively. This highlights its enhanced capability to discern subtle quality variations.
In contrast, while DPO often achieves the highest accuracy on Easy cases where style biases can guide the model, it exhibits a significant performance drop on Normal and Hard ones. For instance, on Mistral-7B-Base model, DPO's accuracy plummets from 89.8\% on Easy cases to just 18.7\% on Hard ones.

\begin{figure*}[!t]
  \centering
  \vskip -1em
  \begin{subfigure}[b]{0.26\linewidth}
    \includegraphics[width=\linewidth]{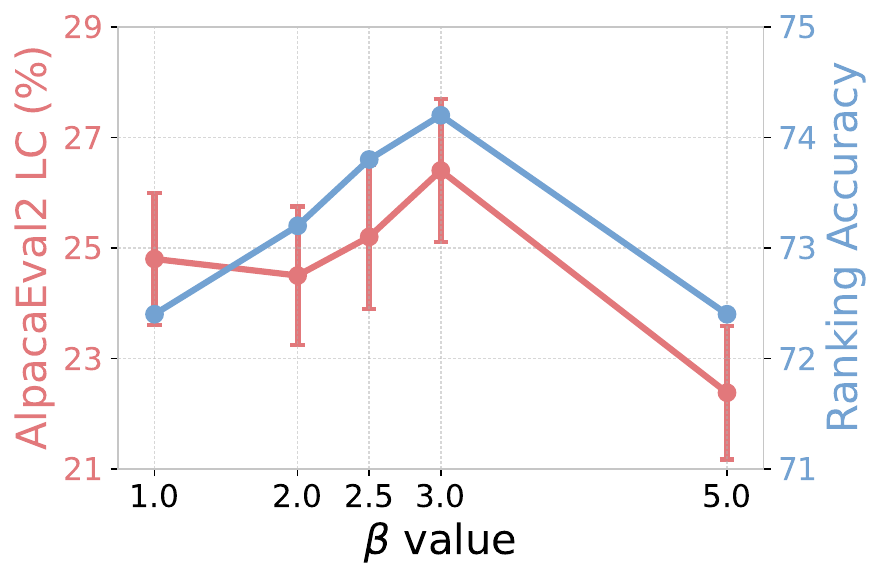}
    \caption{Performance w/ different $\beta$}
    \label{fig:beta-main}
  \end{subfigure}
\hspace{0.02\linewidth} 
  \begin{subfigure}[b]{0.26\linewidth}
    \includegraphics[width=\linewidth]{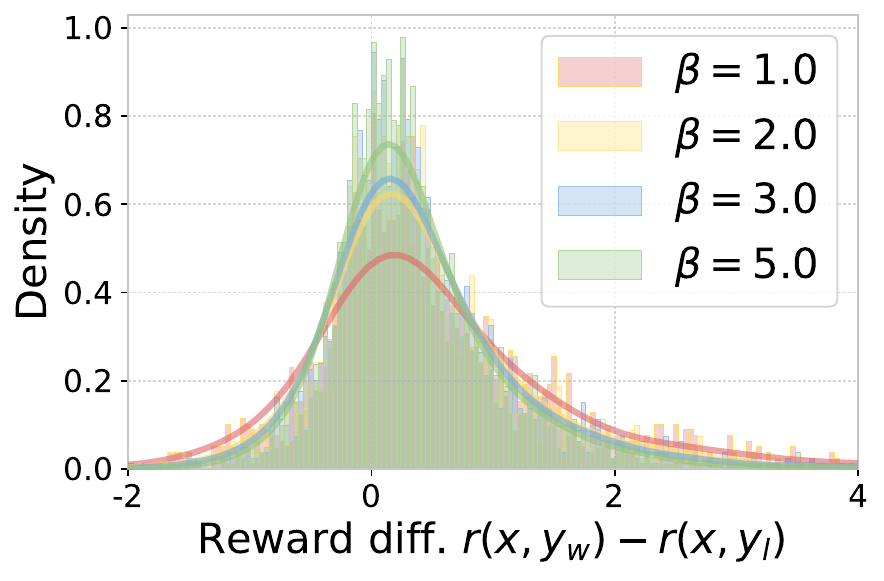}
    \caption{Prob. diff. distribution.}
    \label{fig:beta-reward-diff}
  \end{subfigure}
\hspace{0.02\linewidth} 
  \begin{subfigure}[b]{0.26\linewidth}
    \includegraphics[width=\linewidth]{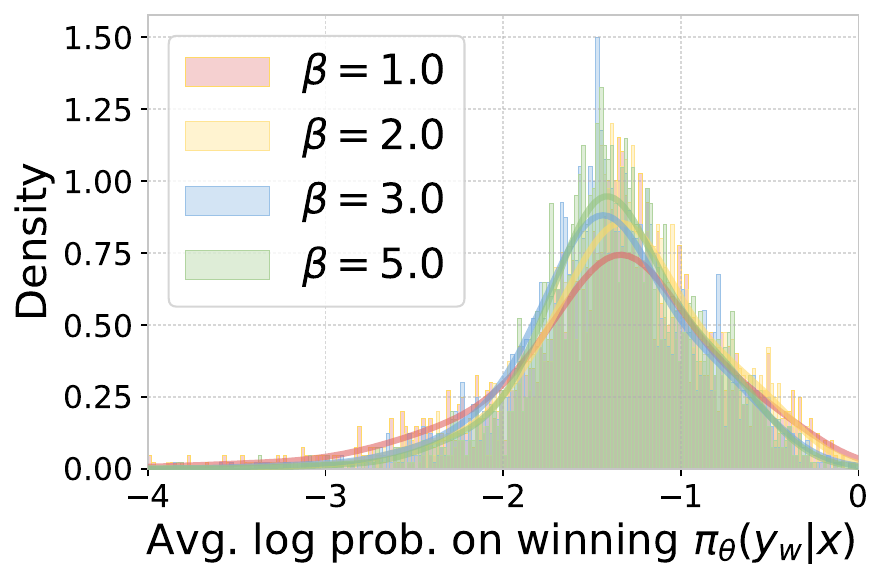}
    \caption{Log prob. distribution.}
    \label{fig:beta-reward-distribution}
  \end{subfigure}
  \caption{Ablation studies of the $\beta$. (a) Ranking accuracy and AlpacaEval2 LC win rate under different $\beta$ values. (b) Reward difference distribution under different $\beta$ values. (c) Log likelihood distribution on chosen responses under different $\beta$ values.}
  \label{fig:main}
  \vskip -1em
\end{figure*}

This discrepancy aligns perfectly with our gradient analysis, which identifies DPO's tendency to overfit to simple patterns while underfitting to complex, misranked samples.
Although SFT models sometimes perform well, especially on instruct-tuned setups, this strength does not reliably carry over to downstream tasks. 
In contrast, AMaPO's robust preference learning translates directly into superior alignment. 
As shown in Table~\ref{main-table}, AMaPO consistently leads on AlpacaEval 2 and MT-Bench.
Notably, it achieves a length-controlled (LC) win rate up to 4.4 points higher than SimPO on Llama3-8B-Base and demonstrates strong performance gains regardless of the evaluation metric.
These findings confirm that AMaPO effectively enhances core ranking capabilities, and this fundamental improvement successfully generalizes to diverse downstream applications. \footnote{For more results like further downstream task results, batch size ablation, generation length, and cases, see Appendices \ref{sec:Downstream_tasks}, \ref{app: Ablation Study of Oracle Margin Estimation}, \ref{app: Reward Distribution of Different Models}, \ref{sec:Length_Analysis}, and \ref{sec:case_study}.}

\setlength{\tabcolsep}{1mm}
\begin{table}[h]
    \centering
    \small
    \caption{Overfitting-underfitting Problem over ultrafeedback dataset following \cite{lin2024limited, hong2025robustness}. Our AMaPO can achieve promising ranking accuracy across in-domain and out-of-distribution settings.}
    \vspace{-3pt}
  \label{table:overfitting-underfitting}
  \renewcommand{\arraystretch}{0.8}
  \begin{tabular}{llcccc}
    \toprule[1pt]
    & \textbf{Method} & $\mathcal{D}_\mathrm{ID}$ & $\mathcal{D}_{\text {$\sim$Prompt}}$ & $\mathcal{D}_{\text {$\sim$Response}}$ & $\mathcal{D}_{\text {$\sim$Mutual}}$\\
    \midrule[0.5pt]
    \multirow{3}{*}{\makecell[c]{Mistral-7B \\ Base}}
    & DPO & 69.66 & 83.01 & 68.41 & 67.35 \\
    & SimPO & 78.02 & 91.05 & 76.08 & 75.94 \\
    & AMaPO & \textbf{79.26} & \textbf{92.46} & \textbf{83.62} & \textbf{82.97} \\
     \midrule[0.8pt]
    \multirow{3}{*}{\makecell[c]{Llama3-8B \\ Base}}
    & DPO & 64.73 & 76.2 & 61.29 & 60.25 \\
    & SimPO & \textbf{78.1} & 91.14 & \textbf{78.6} & 76.3 \\
    & AMaPO & 77.33 & \textbf{91.41} & 77.87 & \textbf{76.91}\\
     \midrule[0.8pt]
    \multirow{3}{*}{\makecell[c]{Mistral-7B \\ Instruct}}
    & DPO & 82.19 & 86.83 & 75.67 & 72.89 \\
    & SimPO & 81.39 & \textbf{95.17} & 86.31 & 85.64 \\
    & AMaPO & \textbf{82.85} & 94.88 & \textbf{91.74} & \textbf{90.91} \\
    \midrule
    \multirow{3}{*}{\makecell[c]{Llama3-8B \\ Instruct}}
    & DPO & 79.23 & 87.34 & 67.94 & 66.16 \\
    & SimPO & \textbf{84.45} & 95.68 & 93.46 & 93.16 \\
    & AMaPO & 84.04 & \textbf{96.5} & \textbf{94.16} & \textbf{94.26} \\
    \bottomrule[1pt]
  \end{tabular}
  \vspace{-3pt}
\end{table}

\paragraph{Results on the Overfitting-Underfitting Problem.}
To further validate our theoretical claims regarding the overfitting-underfitting dilemma, we evaluate model performance across four generalization scenarios. 
As shown in Table~\ref{table:overfitting-underfitting}, the results strongly support our analysis and demonstrate the effectiveness of AMaPO. 
DPO consistently performs the worst across both in-distribution and OOD settings, confirming its dual vulnerability. 
For example, on the Llama3-8B-Base setup, DPO's accuracy is not only the lowest in-distribution but also fails to generalize, lagging behind AMaPO by over 15 points on Prompt OOD. 
This aligns with our theory that DPO underfits on challenging OOD samples.
SimPO, while achieving high in-distribution accuracy, suggesting a tendency to overfit to seen data, does not generalize as effectively as AMaPO due to its static margin. 
For instance, on Mistral-7B-Base, SimPO's performance drops by 15 points from Prompt OOD to Mutual OOD.
In contrast, AMaPO resolves this tension: it matches or exceeds SimPO's in-distribution performance while achieving the best results across all OOD settings. 
This robust generalization confirms that our adaptive margin successfully mitigates the overfitting-underfitting problem we identified.

\setlength{\tabcolsep}{0.5mm}

\begin{table}[h]
    \centering
    \caption{Ablation studies under Llama3-8B setups. We ablate the key design of AMaPO and explore variants of the adaptive margin.}
    \label{Ablation_studies}
    \small
    \renewcommand{\arraystretch}{0.85}
    \begin{tabular}{lcccccc}
        \toprule
        \multirow{3}{*}{\textbf{Method}} & \multicolumn{3}{c}{\textbf{Llama3-8B-Base}} & \multicolumn{3}{c}{\textbf{Llama3-8B-Instruct}} \\
        \cmidrule(r){2-4} \cmidrule(r){5-7}
        & \multicolumn{2}{c}{\textbf{\scriptsize AlpacaEval 2}} & \textbf{\scriptsize MT-Bench} & \multicolumn{2}{c}{\textbf{\scriptsize AlpacaEval 2}} & \textbf{\scriptsize MT-Bench}\\
        \cmidrule(r){2-3} \cmidrule(r){4-4} \cmidrule(r){5-6} \cmidrule(r){7-7}
        & \textbf{\scriptsize LC} \scriptsize(\%) & \textbf{\scriptsize WR} \scriptsize (\%) & \textbf{\scriptsize GPT-4 Turbo} & \textbf{\scriptsize LC} \scriptsize (\%) & \textbf{\scriptsize WR} \scriptsize (\%) & \textbf{\scriptsize GPT-4 Turbo}\\
        \midrule
        DPO & 18.2 & 15.5 & 6.5 & 40.3 & 37.9 & 7.0\\
        SimPO & 22.0 & 20.3 & \textbf{6.6} & 44.7 & 40.5 & 7.0\\
        \midrule
        AMaPO & \textbf{26.3} & \textbf{21.4} & 6.5 & 46.1 & 41.3 & \textbf{7.2}\\
        \midrule 
        w/o Z-norm & 24.8 & 20.4 & 6.3 & 44.2 & 39.6 & 7.0\\ 
        w/o exp & 24.0 & 17.6 & 6.1 & 47.1 & 42.9 & 7.0\\ 
        w/o adaptive & 20.7 & 16.4 & 6.3 & 47.2 & 42.5 & 7.0\\ 
        w/o zero & 22.3 & 21.1 & \textbf{6.6} & \textbf{48.6} & \textbf{45.0} & 6.7 \\

        \bottomrule
    \end{tabular}
\end{table}

\subsection{Ablations and Further Analysis}
\label{sec:analysis}
This section delves deeper into the components and behavior of AMaPO. We begin by ablating the modules, such as Z-normalization and exponential scaling, to confirm their necessity. Our main analysis then focuses on the hyperparameter $\beta$, which controls the strength of the adaptive margin. 

\paragraph{Importance of the components of AMaPO.}
Due to the significant computational resources required, we conduct this ablation study on the Llama3-8B-Base and Llama3-8B-Instruct models, evaluating performance on key downstream benchmarks. Our results in Table~\ref{Ablation_studies} shows that the complete AMaPO design, which integrates Z-normalization (can also been shown as an alternative of linear scaling functions), exponential scaling, and a zero margin for correctly ranked samples, is critical for achieving robust and well-rounded performance. While most ablated variants still outperform the DPO baseline, removing any single component results in a notable performance degradation. This is particularly evident on the Llama3-8B-Base model, where removing exponential scaling and Z-normalization (`w/o adaptive`) degrades the AlpacaEval 2 LC and WR to 20.7\% and 16.7\%, below the full AMaPO's 26.3\% and 21.4\%. Similarly, removing any components of AMaPO on the Llama3-8B-Instruct model causes the score on MT-Bench to drop from AMaPO's 7.2 to about 7.0.  Thus, the synergistic combination of all components in AMaPO is essential for a better adaptive margin and ensuring consistently superior performance across diverse evaluation settings.

\paragraph{Influence of $\beta$.}
To investigate the influence of our sole hyperparameter, $\beta$, we trained AMaPO on the UltraFeedback dataset using the Llama3-8B-Base model with varying $\beta$ values. As shown in Figure~\ref{fig:beta-main}, our analysis on the AlpacaEval 2 and UltraFeedback test sets reveals a clear inverted U-shaped relationship between $\beta$ and performance. Both ranking accuracy and the LC win rate initially increase with $\beta$ before declining, indicating that an optimal regularization strength is empirically around $\beta=3$. The underlying reason for this trend is revealed by analyzing the probability distributions on the UltraFeedback test set (Figures~\ref{fig:beta-reward-diff} and \ref{fig:beta-reward-distribution}). As $\beta$ increases, the distribution of the probability margin, $\pi_\theta(y_w|x) - \pi_\theta(y_l|x)$, sharpens significantly, but the value at the peak of the likelihood of winning responses, $\pi_\theta(y_w|x)$, decreases. This reveals a critical trade-off: while a stronger regularization (a higher $\beta$) initially forces the model to more accurately fit our adaptive margin, leading to performance gains, an excessively high $\beta$ causes the likelihood distribution to become overly peaked, resulting in model degeneration and a subsequent drop in performance.

\section{Conclusion}
In this work, we presented a novel analysis of offline preference optimization and proposed a new algorithm, AMaPO, to address the dilemma we identified. Through a unified margin-based framework, our gradient analysis revealed that methods like DPO suffer from an overfitting-underfitting dilemma, inefficiently learning from preference data. AMaPO rectifies this by introducing an instance-wise adaptive margin that intelligently prioritizes difficult, misranked samples while ignoring those already learned. Extensive experiments show that AMaPO has a better ranking accuracy, leading to a better performance on downstream tasks. By providing both a more effective alignment algorithm and a sharper analytical lens, this work paves the way for more principled advances in preference optimization. \footnote{A detailed discussion of limitations can be found in Appendix~\ref{sec:limitation}.}

\section*{Acknowledgments}
This work was supported in part by the National Natural Science Foundation of China (No. 62272330, and No. U24A20328), and in part by the Science Fund for Creative Research Groups of Sichuan Province Natural Science Foundation (No. 2024NSFTD0035).

\bigskip

\bibliography{aaai2026}
\appendix
\newpage

\section{Limitations and Future Work}
\label{sec:limitation}
While our work introduces a principled approach to preference optimization and demonstrates its effectiveness, we identify several promising avenues for future research that arise from the scope and limitations of our study.

\paragraph{Scaling.}
Our experiments were conducted on models up to the 8B parameter scale. A crucial next step is to investigate the performance and dynamics of AMaPO on much larger models (e.g., 70B and beyond), where phenomena like overfitting and underfitting may manifest differently. 

\paragraph{Generality of the Oracle Margin Estimator.}
We propose using the batch-wise mean implicit margin as a practical and effective proxy for the inaccessible Oracle Ranking Margin. However, the optimality of this non-parametric estimator may not be universal. Its performance could be influenced by factors such as the quality and diversity of the initial SFT model or the noise level and distribution of the preference dataset. Future research could explore more sophisticated estimators for the oracle margin, potentially involving lightweight parametric models or meta-learning approaches, to further enhance the adaptability of the margin.

\paragraph{Alternatives of scaling functions.}
In this work, the scaling function serves to appropriately amplify the gradient magnitude of hard, incorrectly ranked samples. AMaPO exclusively consider the widely-used exponential function and Z-normalization as the nonlinear and linear scaling functions, respectively, which elegantly transform our additive, log-space margin into the geometric mean of the PPL ratio within the batch. Although our ablation studies demonstrate that the popular linear and non-linear function do work, a comprehensive exploration of better scaling functions would further improve the margin-based preference optimization algorithms.

\paragraph{Scope of the Gradient Analysis.}
Our theoretical motivation is grounded in a gradient-level analysis that examines the instantaneous effect of the margin on the learning signal. This static snapshot is powerful for algorithm design but does not constitute a full analysis of the dynamic training process over time. A compelling direction for future work would be to study the entire training trajectory, exploring how AMaPO's adaptive margin influences the loss landscape, convergence properties, and the evolution of the model's internal representations throughout the optimization process.

\section{Gradient Analysis} 
In this section, we employ the unified framework to analyze other methods in detail.

\subsection{Baseline Methods} \label{sec:baseline_methods}
We also provides a detailed introduction to other state-of-the-art baselines of offline preference optimization, with an emphasis on the design of the margin in their learning objectives.


\textbf{IPO} Identity Preference Optimization \cite{azar2024general} minimizes a squared loss regression
problem by defining an alternative reward function, avoiding unstable RL training. IPO involves one hyperparameter, $\beta$, to adjust the reward margin.
\begin{equation*} 
\mathcal{L}_{\mathrm{IPO}}=-
\left[ {\left(\log \tfrac{\pi_\theta\left(y_w \mid x\right)}{\pi_\theta\left(y_l \mid x\right)} - \log \tfrac{\pi_{\mathrm{ref}}\left(y_w \mid x\right)}{\pi_{\mathrm{ref}}\left(y_l \mid x\right)}-\tfrac{1}{2\beta} \right)}^2 \right].
\end{equation*}

\textbf{SLiC} Sequence Likelihood Calibration \cite{zhao2023slic} directly uses log-likelihood as a ranking loss and includes a SFT objective to prevent degeneration. SLiC designs a hyperparameter $\delta$ as the margin for the ranking loss, where the ranking loss decreases with the ranking margin increases.
\begin{equation*} 
\mathcal{L}_{\mathrm{SLiC}}=
\mathrm{max}\left(0, \delta-\log \tfrac{\pi_\theta\left(y_w \mid x\right)}{\pi_\theta\left(y_l \mid x\right)}\right) - \lambda \log \pi_\theta(y_w \mid x).
\end{equation*}

It can also be interpreted as an hinge loss variation of DPO \cite{liu2023statistical} as follows:
\begin{equation*}
    \mathcal{L}_{\mathrm{SLiC}} = \mathrm{max}\left\{0, 1-\tau(\log \tfrac{\pi_\theta\left(y_w \mid x\right)}{\pi_\theta\left(y_l \mid x\right)} - \log\tfrac{\pi_{\mathrm{ref}}\left(y_w \mid x\right)}{\pi_{\mathrm{ref}}\left(y_l \mid x\right)})\right\},
\end{equation*}
where $\tau=\frac{1}{\delta}$ denotes the true label in hinge loss.

\textbf{CPO} Contrastive Preference Optimization \cite{xu2024contrastive} uses log-likelihood as the reward and introduces an explicit regularization with Supervised Fine-Tuning (SFT) objective rather than the reward margin in BT-model, while eliminating the need for the reference policy.
\begin{equation*} 
\mathcal{L}_{\mathrm{CPO}}=
-\log \sigma\left(\beta\log \tfrac{\pi_\theta\left(y_w \mid x\right)}{\pi_\theta\left(y_l \mid x\right)}\right) - \lambda \log \pi_\theta(y_w \mid x).
\end{equation*}

\textbf{ODPO} Direct Preference Optimization with an offset \cite{amini2024direct} incorporates additional ground-truth reward values from external reward model and employs the reward values difference $\Delta_r$ as an instance-wise margin.
\begin{align*} 
\mathcal{L}_{\mathrm{ODPO}}=-
\bigg[ \log \sigma\bigg( & \beta\log \tfrac{\pi_\theta\left(y_w \mid x\right)}{\pi_\theta\left(y_l \mid x\right)} \\
& - \beta\log\tfrac{\pi_{\mathrm{ref}}\left(y_w \mid x\right)}{\pi_{\mathrm{ref}}\left(y_l \mid x\right)} \bigg) - \Delta_r\bigg].
\end{align*}


\textbf{FocalPO} FocalPO \cite{liu2025focalpo} downweighs misranked preference pairs and prioritizes enhancing the model’s understanding of pairs that it can already rank correctly.
\begin{align*} 
\mathcal{L}_{\mathrm{FocalPO}}= -\bigg[ p^{\gamma}\left(y_w \succ y_l \mid x\right) \log p\left(y_w \succ y_l \mid x\right) \bigg].
\end{align*}
where $p\left(y_w \succ y_l | x\right) = \sigma (\beta\log \frac{\pi_\theta\left(y_w \mid x\right)}{\pi_\theta\left(y_l \mid x\right)} - \beta\log \frac{\pi_\mathrm{ref}\left(y_w \mid x\right)}{\pi_\mathrm{ref}\left(y_l \mid x\right)} )$.

\textbf{KTO} Kahneman-Tversky Optimization \cite{ethayarajh2024ktomodelalignmentprospect} learns from non-paired preference data to adapt to a board usage.
\begin{align*} 
\mathcal{L}_{\mathrm{KTO}}= &- \lambda_w \sigma \left( \beta \log \tfrac{\pi_\theta\left(y_w \mid x\right)}{\pi_\mathrm{ref}\left(y_w \mid x\right)} - z_\mathrm{ref}\right) \\ 
& + \lambda_l\sigma\left( z_\mathrm{ref} - \beta \log \tfrac{\pi_\theta\left(y_l \mid x\right)}{\pi_\mathrm{ref}\left(y_l \mid x\right)} \right),
\end{align*}
where $z_\mathrm{ref}=\mathbb{E}_{\left(x, y\right) \sim \mathcal{D}}\left[ \beta \mathrm{KL}(\pi_\theta(y \mid x)||\pi_\mathrm{ref}(y \mid x)) \right]$.

\textbf{$\alpha$-DPO} Adaptive Reward Margin for Direct Preference Optimization \cite{wu2024alpha} interpolates between policy-driven specialization and uniform exploration while enabling instance-adaptive reward margins.
\begin{align*} 
\mathcal{L}_{\mathrm{\alpha-DPO}}=-
\bigg[ \log \sigma\bigg( & u(x,y_w,y_l) \\ 
& - \mathrm{sg}\big[\gamma + \alpha M^*(x,y_w,y_l) \big] \bigg) \bigg].
\end{align*}
where $u(x,y_w,y_l)=\frac{\beta}{|y_w|}\log \pi_\theta(y_w|x)-\frac{\beta}{|y_l|}\log \pi_\theta(y_l|x)$, $M(x,y_w,y_l)=\beta \log \frac{\pi_\theta(y_w|x)\pi_{\mathrm{ref}}(y_l|x)}{\pi_{\mathrm{ref}}(y_w|x)\pi_\theta(y_l|x)}$, $M^*$ represents applying Z-normalization to $M$ across training dataset, $\gamma$ is a tunable constant margin and sg represent stop gradient.

Under our unified margin framework Eq. \ref{unified_framework}, 
they are shown as Table \ref{app:unify}.

\setlength{\tabcolsep}{1mm}
\begin{table*}[h]
\small
  \caption{The state-of-the-art baselines for offline preference optimization objectives under our unified framework. Constants satisfy $\beta,\gamma,\delta,\Delta_r,\lambda_w,\lambda_l>0$, and where $ c_\mathrm{ref}^w=\log \pi_{\mathrm{ref}}(y_l|x)$, $ c_\mathrm{ref}^l= \log \pi_{\mathrm{ref}}(y_l|x)$, $c_\mathrm{ref}= c_\mathrm{ref}^w - c_\mathrm{ref}^l$}.
  \label{app:unify}
  \centering 
  \begin{tabular}{lccccc}
    \toprule
    & $m(a)$ & $h_w(a)$ & $h_l(a)$ & $\gamma$ & $\Lambda(a)$ \\
    \toprule
    DPO \cite{rafailov2023direct} & $\log \sigma(a)$ & $\beta a$ & $\beta a$ & $\beta c_\mathrm{ref}$ & - \\
    IPO \cite{azar2024general} & $a^2$ & $a$ & $a$ & $c_\mathrm{ref}+\frac{1}{2\beta}$ & - \\
    SLiC \cite{zhao2023slic} & $\mathrm{max}(0, 1-\tau\cdot a)$ & $a$ & $a$ & $c_\mathrm{ref}$ & - \\
    CPO \cite{xu2024contrastive} & $\log \sigma(a)$ & $\beta a$ & $\beta a$ & - & $\lambda a$ \\
    ODPO \cite{amini2024direct} & $\log \sigma(a)$ & $\beta a$ & $\beta a$ & $\beta c_\mathrm{ref}+\Delta_r$ & - \\
    KTO \cite{ethayarajh2024ktomodelalignmentprospect} & $a$ & $\lambda_w\sigma(\beta a - \beta c_\mathrm{ref}^w - z_{\mathrm{ref}})$ & $\lambda_l\sigma(\beta c_\mathrm{ref}^l - \beta a + z_{\mathrm{ref}})$ & - & - \\
    SimPO \cite{meng2024simpo} & $\log \sigma(a)$ & $\frac{\beta}{|y_w|} a$ & $\frac{\beta}{|y_l|} a$ & $\gamma$ & - \\
    FocalPO \cite{liu2025focalpo} & $\sigma^{\gamma}(a) \log \sigma(a)$ & $\beta a$ & $\beta a$ & $\beta c_\mathrm{ref}$ & - \\
    $\alpha$-DPO \cite{wu2024alpha} & $\log \sigma(a)$ & $\frac{\beta}{|y_w|} a$ & $\frac{\beta}{|y_l|} a$ & $\gamma+\alpha M^*$ & - \\
    \bottomrule
  \end{tabular}
  \vspace{-10pt}
\end{table*}

\subsection{Gradient Analysis}\label{Gradient Analysis}
We further analyze the gradients of the aforementioned objectives under our unified framework Eq. \ref{eq:gradient_full} (except DPO and SimPO) to understand the influence of the margin. For notational simplicity, we sometimes omit the $\beta$ in the following derivations, and $\pi_\theta(x,y_w,y_l) = \log \pi_\theta(y_w|x)-\log \pi_\theta(y_l|x)$.

\textbf{IPO} The gradient magnitude $d_{\boldsymbol{\theta}}$ is a monotonically increasing function. 
\begin{align*}
\nabla_{\boldsymbol{\theta}} \mathcal{L}_{\mathrm{IPO}}\left(\theta\right) = & -\underbrace{2(\pi_\theta(x,y_w,y_l)-(c_\mathrm{ref}+\tfrac{1}{2\beta}))}_{\text{gradient magnitude }d_{\boldsymbol{\theta}}} \\
&(\nabla\log \pi_w - \nabla\log \pi_l).
\end{align*}
The margin, $\gamma = c_\mathrm{ref}+\frac{1}{2\beta}$, derived from the reference model and calibrated reward margin, might result in overfitting and underfitting, similar to DPO.

\textbf{SLiC} Considering $1-\tau(\log \frac{\pi_\theta\left(y_w \mid x\right)}{\pi_\theta\left(y_l \mid x\right)} - \log\frac{\pi_{\mathrm{ref}}\left(y_w \mid x\right)}{\pi_{\mathrm{ref}}\left(y_l \mid x\right)})>0$, i.e., when the gradient exists. The gradient magnitude $d_{\boldsymbol{\theta}}$ is a monotonically increasing function. The margin, $\gamma = c_\mathrm{ref}$, derived from the reference model, might also result in underfitting.
\begin{align*}
\nabla_{\boldsymbol{\theta}} \mathcal{L}_{\mathrm{SLiC}}\left(\theta\right) = &-\underbrace{\tau(\pi_\theta(x,y_w,y_l)-c_\mathrm{ref})}_{\text{gradient magnitude }d_{\boldsymbol{\theta}}}\\
&(\nabla\log \pi_w - \nabla\log \pi_l).
\end{align*}

\textbf{CPO} The first gradient magnitude $d^w_{\boldsymbol{\theta}}$ is  monotonically increasing, where the explicit regularization with SFT $\lambda>0$ manipulates the gradient magnitude. However, the margin, $\gamma = 0$, might result in underfitting. And $d^l_{\boldsymbol{\theta}}$ likewise.

\begin{align*}
\nabla_{\boldsymbol{\theta}} \mathcal{L}_{\mathrm{CPO}}\left(\theta\right) =& \underbrace{(\sigma(\pi_\theta(x,y_w,y_l))-1-\lambda)}_{\text{gradient magnitude }d^w_{\boldsymbol{\theta}}}\nabla\log \pi_w \\ 
& + \underbrace{(1-\sigma(\pi_\theta(x,y_w,y_l)))}_{\text{gradient magnitude }d^l_{\boldsymbol{\theta}}}\nabla\log \pi_l,
\end{align*}
where $\pi_\theta(x,y_w,y_l) = \log \pi_\theta(y_w|x)-\log \pi_\theta(y_l|x)$.

\textbf{ODPO} The gradient magnitude $d_{\boldsymbol{\theta}}$ is a monotonically decreasing function. The margin, $\gamma = c_\mathrm{ref}+\Delta_r$ derived from the reference model and annotated reward margin, might result in overfitting and underfitting, similar to DPO.
\begin{align*}
\nabla_{\boldsymbol{\theta}} \mathcal{L}_{\mathrm{ODPO}}\left(\theta\right) = & -\underbrace{(1-\sigma(\pi_\theta(x,y_w,y_l) - ( c_\mathrm{ref}+\Delta_r))}_{\text{gradient magnitude }d_{\boldsymbol{\theta}}} \\
& (\nabla\log \pi_w - \nabla\log \pi_l).
\end{align*}

\textbf{FocalPO} The gradient magnitude $d_{\boldsymbol{\theta}}$ is monotonically decreasing for $\sigma(u) < e^{-\frac{1}{\gamma}}$, and monotonically increasing for $\sigma(u) > e^{-\frac{1}{\gamma}}$. When applied to the uncommon case of a non-monotonic $d_{\boldsymbol{\theta}}$, our theoretical framework does not readily yield a definitive conclusion, necessitating further analysis.

\begin{align*}
\nabla_{\boldsymbol{\theta}} \mathcal{L}_{\mathrm{FocalPO}}\left(\theta\right) = & -\underbrace{\sigma(u)^\gamma(1-\sigma(u))(\gamma \log \sigma(u)+1)}_{\text{gradient magnitude }d_{\boldsymbol{\theta}}} \\
& (\nabla\log \pi_w - \nabla\log \pi_l).
\end{align*}
where $u = \pi_\theta(x,y_w,y_l) - c_\mathrm{ref}$

\textbf{KTO} The gradient magnitude $d^w_{\boldsymbol{\theta}}$ is monotonically increasing when $\beta \log \pi_\theta(y_w|x) -(\beta c_\mathrm{ref}^w + z_{\mathrm{ref}})<0$, and monotonically decreasing when $\beta \log \pi_\theta(y_w|x) -(\beta c_\mathrm{ref}^w + z_{\mathrm{ref}})>0$. Therefore, we can view $\beta c_\mathrm{ref}^w + z_{\mathrm{ref}}$ as the margin of KTO, which is derived from the reference model and KL divergence. However, the margin might result in overfitting and underfitting, similar to DPO. And $d^l_{\boldsymbol{\theta}}$ likewise.

\begin{align*}
\nabla_{\boldsymbol{\theta}} \mathcal{L}_{\mathrm{KTO}}\left(\theta\right) = &-\underbrace{\lambda_w\sigma_w(1-\sigma_w)}_{\text{gradient magnitude }d^w_{\boldsymbol{\theta}}}\nabla\log \pi_w \\
& + \underbrace{\lambda_l\sigma_l(1-\sigma_l)}_{\text{gradient magnitude }d^l_{\boldsymbol{\theta}}}\nabla\log \pi_l,
\end{align*}
where $\sigma_w=\sigma(\beta \log \pi_\theta(y_w|x) -(\beta c_\mathrm{ref}^w + z_{\mathrm{ref}}))$ and $\sigma_l=\sigma(\beta c_\mathrm{ref}^l + z_{\mathrm{ref}} - \beta \log \pi_\theta(y_l|x))$.

\textbf{$\alpha$-DPO} The gradient magnitude $d_{\boldsymbol{\theta}}$ is a monotonically decreasing function. The margin, $\gamma+\alpha M^*$ derived from policy-driven specialization and a tunable target margin, might result in overfitting and underfitting, similar to DPO, while introducing multiple tunable hyperparameters.
\begin{align*}
\nabla_{\boldsymbol{\theta}} \mathcal{L}_{\alpha-\mathrm{DPO}}\left(\theta\right) = & -\underbrace{(1-\sigma(\pi_\theta(x,y_w,y_l) - ( \gamma+\alpha M^*))}_{\text{gradient magnitude}d_{\boldsymbol{\theta}}} \\
& (\nabla\log \pi_w - \nabla\log \pi_l).
\end{align*}

\section{Further Theoretical Analysis}

\subsection{Analysis towards Oracle Ranking Margin}\label{app:analysis_oracle_margin}
We have empirically claimed that Oracle Ranking Margin $\gamma^*$ is an instance-specific, non-negative threshold, as stated in our Definition~\ref{def:oracle_margin}:
\begin{equation*}
    \gamma^* = r^*(x,y_w) - r^*(x,y_l)
\end{equation*}
where $r^*(\cdot)$ represents the optimal reward function.

Here, we theoretically verify our claim regarding the properties of $\gamma^*$ based on the theoretical analysis and proof in \cite{rafailov2023direct, liu2024understanding}:
\begin{lemma}
\label{reward_function}
    All reward classes under the BT model can be represented with the reparameterization $r(x,y)=\beta \log \pi_\theta(y|x)$ for some policy model $\pi_\theta(y|x)$.
\end{lemma}

Therefore, the optimal reward function $r^*(\cdot)$ can be reparameterized with a certain policy model $\pi^*_\theta(y|x)$ as follows:
\begin{equation}
    \label{oracle_margin_reparameterization}
    \gamma^*(x,y_w,y_l) = \beta \log \pi^*_\theta(x|y_w) - \beta \log \pi^*_\theta(x|y_l)
\end{equation}
where $\pi^*_\theta(x|y)$ varies with $x$ and $y$, and thus $\gamma^*$ should be an instance-specific threshold $\gamma^*(x,y_w,y_l)$.

Meanwhile, the reparameterization of $\gamma^*$ as presented in Eq~\ref{oracle_margin_reparameterization} lends theoretical backing to our estimation method for $\gamma^*$ in Eq~\ref{AMaPO_margin}, which leverages the mean log probability of responses within the current training batch to estimate $\gamma^*$.

\subsection{Derivations for Margin Scaling}\label{Derivations for Margin Scaling}

In our method, we introduced the exponential scaling function, $h_\gamma(\gamma)$, with the motivation that it better represents the quality gap between responses by leveraging perplexity (PPL). This appendix provides the formal mathematical justification for this choice. We will show that the exponential form has a compelling and intuitive interpretation: it is directly related to the geometric mean of the PPL ratios between the losing and winning responses, adjusted by the sample's specific difficulty.

For $\gamma > 0$ in Eq. \ref{margin_scaling}, we have
\begin{equation*}
    h_\gamma(\gamma) = \beta e^{\gamma} = \beta e^{\alpha \mu_r} = \beta {(e^{\mu_r})}^{\alpha}, \\
\end{equation*}
where $\alpha = \frac{\mu_r-r_{\pi_\theta}(x,y_w,y_l)}{\sigma_r}$ is the Z-normalization coefficient for $r_{\pi_\theta}(x,y_w,y_l)$, and $\beta$ is the scaling constant. Therefore, we can focus on $e^{\mu_r}$, where
\begin{align*}
    e^{\mu_\gamma}=&\mathrm{exp}({\frac{1}{|B|}\sum_{i=1}^{|B|}|y^i_w| \log \pi_\theta(y^i_w|x^i)-|y^i_l|\log \pi_\theta(y^i_l|x)}) \\
    =&(\prod_{i=1}^{|B|}\mathrm{exp}(|y^i_w| \log \pi_\theta(y^i_w|x^i)-|y^i_l|\log \pi_\theta(y^i_l|x)))^\frac{1}{|B|} \\
    =&(\prod_{i=1}^{|B|} \frac{\mathrm{exp}(-|y^i_l|\log \pi_\theta(y^i_l|x))}{\mathrm{exp}(-|y^i_w| \log \pi_\theta(y^i_w|x^i))})^\frac{1}{|B|} \\
    =&(\prod_{i=1}^{|B|} \frac{\mathrm{PPL}(y^i_l|x)}{\mathrm{PPL}(y^i_w|x)})^\frac{1}{|B|} \\
    =&G(\frac{\mathrm{PPL}(y_l|x)}{\mathrm{PPL}(y_w|x)}),
\end{align*}
which means that the exponential scaling function represents the geometric mean of the ratios of PPL for losing and winning responses within the training batch $B$, capturing their quality difference.

Finally, we have an intuitive understanding of Margin Scaling, which represents the quality gap between responses within the training batch.
\begin{equation*}
    h_\gamma(\gamma) = \beta G(\frac{\mathrm{PPL}(y_l|x)}{\mathrm{PPL}(y_w|x)})^{\alpha}.
\end{equation*}

\section{Experimental Details} \label{app:Experimental Details}

\subsection{Implementation Details} \label{sec:Implementation_Details}

\textbf{Overfitting-underfitting Problem Setups.}
\label{sec:overfitting_underfitting_setups}
To evaluate the Overfitting-underfitting Problem across offline preference optimization baselines (i.e. DPO and SimPO), we follow the methodology of \cite{lin2024limited, hong2025robustness}, which evaluates ranking accuracy across the following four generalization scenarios based on prompt disjointness and response disjointness\footnote{\url{https://huggingface.co/collections/rm-robustness}}.

Let $\mathcal{X}_\mathrm{train} \subset \mathcal{X}$ and $\mathcal{X}_\mathrm{valid} \subset \mathcal{X}$ be the sets of prompts
used for generating responses in the training and validation
sets respectively. $\mathcal{M}_\mathrm{train} \subset \mathcal{M}$ and $\mathcal{M}_\mathrm{valid} \subset \mathcal{M}$ represent the sets of models
used for generating responses likewise. The corresponding response spaces are: $\mathcal{Y}_{\text {train }}(x)=\bigcup_{M_k \in \mathcal{M}_{\text {train}}} \mathcal{Y}_k(x)$ and $\mathcal{Y}_{\text {valid }}(x)=\bigcup_{M_k \in \mathcal{M}_{\text {valid}}} \mathcal{Y}_k(x)$. Therefore, the training set for $\pi_\theta$ can be defined as follows:
\begin{equation*}
    \mathcal{D}_{\text {train }}=\left\{\left(x, y_w, y_l\right) \mid\left(y_w, y_l\right) \sim \mathcal{Y}_{\text {train }}(x), x \in \mathcal{X}_{\text {train }}\right\}.
\end{equation*}

\textit{In-Distribution ($\mathcal{D}_{\text {ID}}$). } $\mathcal{D}_{\text {ID}}$ samples from the seen prompt set $\mathcal{X}_{\text{train}}$ and with the seen response model set $\mathcal{M}_\mathrm{train}$.
\begin{equation*}
    \mathcal{D}_{\text {ID}}=\left\{\left(x, y_w, y_l\right) \mid\left(y_w, y_l\right) \sim \mathcal{Y}_{\text {train }}(x), x \in \mathcal{X}_{\text {train }}\right\}.
\end{equation*}

\textit{Prompt-OOD ($\mathcal{D}_{\text {$\sim$Prompt}}$). } $\mathcal{D}_{\text {$\sim$Prompt}}$ samples from the unseen prompt set $\mathcal{X}_{\text {valid }}$ but with the seen response model set $\mathcal{M}_\mathrm{valid}$. 
\begin{equation*}
    \mathcal{D}_{\text {$\sim$Prompt}}=\left\{\left(x, y_w, y_l\right) \mid\left(y_w, y_l\right) \sim \mathcal{Y}_{\text {train }}(x), x \in \mathcal{X}_{\text {valid }}\right\}.
\end{equation*}

\textit{Response-OOD ($\mathcal{D}_{\text {$\sim$Response}}$).} $\mathcal{D}_{\text {$\sim$Response}}$ samples from the seen prompt set $\mathcal{X}_{\text{train}}$ but with the unseen response model set $\mathcal{M}_\mathrm{valid}$. 
\begin{equation*}
    \mathcal{D}_{\text {$\sim$Response}}=\left\{\left(x, y_w, y_l\right) \mid\left(y_w, y_l\right) \sim \mathcal{Y}_{\text {valid }}(x), x \in \mathcal{X}_{\text {train }}\right\}.
\end{equation*}

\textit{Mutual-OOD ($\mathcal{D}_{\text {$\sim$Mutual}}$).} $\mathcal{D}_{\text {$\sim$Mutual}}$ samples from the unseen prompt set $\mathcal{X}_{\text{valid}}$ and with the unseen response model set $\mathcal{M}_\mathrm{valid}$. 
\begin{equation*}
    \mathcal{D}_{\text {$\sim$Mutual}}=\left\{\left(x, y_w, y_l\right) \mid\left(y_w, y_l\right) \sim \mathcal{Y}_{\text {valid}}(x), x \in \mathcal{X}_{\text {valid }}\right\}.
\end{equation*}

For prompts with multiple responses (i.e. $\mathcal{D}_{\text {$\sim$Prompt}}$, $\mathcal{D}_{\text {$\sim$Response}}$ and $\mathcal{D}_{\text {$\sim$Mutual}}$), we select the highest-scoring one as chosen response $y_w$ and the lowest-scoring one as rejected response $y_l$ to evaluate ranking accuracy. In-distribution ranking accuracy on $\mathcal{D}_{\text {ID}}$ demonstrates the learning capability of $\pi_\theta$, measuring the underfitting. While out-of-distribution ranking accuracy on $\mathcal{D}_{\text {$\sim$Prompt}}$, $\mathcal{D}_{\text {$\sim$Response}}$ and $\mathcal{D}_{\text {$\sim$Mutual}}$ demonstrates the robustness of $\pi_\theta$ across generalization scenarios, measuring the overfitting. Therefore, we define the \textbf{Overfitting-underfitting Problem} as suboptimal in-distribution ranking accuracy and a degradation in out-of-distribution ranking accuracy.

\setlength{\tabcolsep}{2mm}
\begin{table}[h]
  \caption{Searching ranges for hyperparameters across baseline}
  \label{hyperparamater_search}
  \centering 
\renewcommand{\arraystretch}{0.9}
  \begin{tabular}{cc}
    \toprule
    \textbf{Method} & \textbf{Hyperparameter Search Range}\\
    \midrule
    DPO & $\beta \in [0.01,0.05,0.1]$ \\
    \midrule
    IPO & $\tau \in [0.01,0.1,0.5,1.0]$ \\
    \midrule
    \multirow{2}{*}{\makecell[c]{CPO}} & $\lambda=1.0$ \\
    & $\beta \in[0.01,0.05,0.1]$ \\
    \midrule
    \multirow{2}{*}{\makecell[c]{SLiC}} & $\lambda \in [0.1,0.5,1.0,10.0]$ \\
    & $\delta \in[0.1,0.5,1.0,2.0]$ \\
    \midrule
    \multirow{2}{*}{\makecell[c]{KTO}} & $\lambda_l=\lambda_w=1.0$ \\
    & $\beta \in[0.01,0.05,0.1]$ \\
    \midrule
    \multirow{2}{*}{\makecell[c]{SimPO}} & $\beta \in [2.0,2.5]$ \\
    & $\gamma \in[0.3,0.5,1.0,1.2,1.4,1.6]$ \\
    \midrule
    \multirow{2}{*}{\makecell[c]{$\alpha$-DPO}} & $\beta \in [2.5,10.0], \gamma \in [0.1, 0.3, 0.5]$ \\
    & $\alpha \in [1.0e-2, 5.0e-2, 0.1, 0.2]$ \\
    \midrule
    AMaPO & $\beta \in [1.0,2.0,3.0,5.0]$ \\
    \bottomrule
  \end{tabular}
  \normalsize
\end{table}

\textbf{Training Hyperparameters. }
For general hyperparameters, we adhered strictly to the settings used in SimPO. We applied the following hyperparameters: For the SFT stage, we use a learning rate of $2 \times10^{-5}$. For both the SFT and the preference optimization stages, we use a batch size of 128, a maximum sequence length of 2048 (1024 for Mistral-7B-Base setup), and a cosine learning rate schedule with 10\% warmup steps for one epoch, all through the Adam optimizer \cite{kingma2014adam}. We maintain these settings consistently to ensure uniformity and comparability across experiments.

\setlength{\tabcolsep}{1mm}
\begin{table*}[!t]
  \caption{Downstream tasks evaluation results on HuggingFace Open LLM Leaderboard}
  \label{downstream-table}
  \small
  \centering 
  \renewcommand{\arraystretch}{0.95}
  \begin{tabular}{llcccccccccc}
    \toprule
    & \textbf{Method} & \textbf{MMLU-PRO} & \textbf{IFEval} & \textbf{BBH} & \textbf{GPQA} & \textbf{MUSR} & \textbf{MATH} & \textbf{ARC} & \textbf{TruthfulQA} & \textbf{Winograd} & \textbf{Avg. Rank} \\
    \midrule
    \multirow{7}{*}{\makecell[c]{Mistral-7B \\ Base}} & DPO & 26.73 & 10.49 & 43.27 & 28.44 & \textbf{43.65} & 1.36 &  61.26 & 53.06 & 76.80 & 4.1 \\ 
    & SLiC & 26.52 & 12.45 & 42.33 & 27.93 & 33.74 & 1.38  & 55.38 & 48.36 & 77.35 & 5.2 \\
    & IPO & 25.87 & 11.52 & 40.59 & 28.15 & 42.15 & 1.25 & 60.84 & 45.44 & 77.58 & 5.4 \\
    & KTO & 27.51 & 12.03 & 43.66 & 29.45 & 43.17 & 2.34 & 62.37 & \textbf{56.60} & 77.27 & \textbf{2.5} \\
    & CPO & 27.04 & \textbf{13.32} & 42.05 & 28.45 & 42.15 & 2.15 & 57.00 & 47.07 & 76.48 & 4.6 \\
    & SimPO & 27.13 & 10.63 & 42.94 & 29.03 & 39.68 & \textbf{2.49} & \textbf{62.63} & 50.68 & 77.54 & 3.3 \\
    & AMaPO & \textbf{27.52} & 10.45 & \textbf{44.70} & \textbf{31.70} & 42.20 & 1.30 & 62.54 & 52.87 & \textbf{78.37} & 2.8 \\
    \midrule
    \multirow{7}{*}{\makecell[c]{Llama3-8B \\ Base}} & DPO & 31.58 & 33.61 & 47.80 & 32.23 & 40.48 & \textbf{4.53} & 64.42 & 53.48 & 76.80 & 4.0 \\
    & SLiC & 31.11 & 32.37 & 46.53 & 33.29 & 40.55 & 3.92 & 61.43 & 54.95 & 77.27 & 4.8 \\
    & IPO & 30.18 & 31.52 & 46.78 & 32.61 & 39.58 & 4.02 & 62.88 & 54.20 & 72.22 & 6.3 \\
    & KTO & 31.16 & 37.10 & 47.98 & \textbf{33.72} & 40.21 & 4.14 & 63.14 & 55.76 & 76.09 & 3.8 \\
    & CPO & 30.95 & \textbf{38.57} & 47.17 & 33.15 & \textbf{41.59} & 4.25 & 61.69 & 54.29 & 76.16 & 3.9 \\
    & SimPO & \textbf{31.61} & 37.55 & \textbf{48.38} & 33.22 & 40.08 & 4.23 & 65.19 & 59.46 & 76.32 & 2.8 \\
    & AMaPO & 31.38 & 38.47 & 48.17 & 33.12 & 41.12 & 4.13 & \textbf{65.95} & \textbf{60.43} & \textbf{77.54} & \textbf{2.4} \\
    \midrule
    \multirow{7}{*}{\makecell[c]{Mistral-7B \\ Instruct}} & DPO & 26.81 & 22.89 & 45.46 & 28.19 & \textbf{46.43} & 1.89 & \textbf{66.89} & 68.40 & 76.32 & 3.6 \\
    & SLiC & 25.69 & 29.53 & 45.24 & 27.04 & 43.90 & 1.95 & 59.90 & 65.30 & 76.32 & 5.7 \\
    & IPO & 25.75 & 27.85 & 43.81 & 26.61 & 43.55 & 2.02 & 63.31 & 67.36 & 75.85 & 6.0\\
    & KTO & 27.46 & 35.42 & 45.34 & 28.19 & 45.77 & \textbf{2.35} & 65.72 & 68.43 & 75.91 & 3.0\\
    & CPO & 26.85 & 36.81 & 45.01 & 28.15 & 43.28 & 2.28 & 63.23 & 67.38 & 76.80 & 4.3 \\
    & SimPO & 27.10 & 37.52 & 45.70 & 28.04 & 44.71 & 2.19 & \textbf{66.89} & 68.40 & 76.32 & 2.9 \\
    & AMaPO & \textbf{27.85} & \textbf{38.47} & \textbf{46.17} & \textbf{28.36} & 44.92 & 1.97 & 65.79 & \textbf{70.85} & \textbf{77.82} & \textbf{1.9} \\
    \midrule
    \multirow{7}{*}{\makecell[c]{Llama3-8B \\ Instruct}} & DPO & \textbf{35.86} & \textbf{44.57} & 48.31 & 31.04 & 39.02 & 8.23 & 63.99 & 59.01 & 74.66 & 2.3 \\
    & SLiC & 33.25 & 44.01 & 47.55 & 30.52 & 38.10 & \textbf{8.29} & 61.26 & 53.23 & 76.16 & 4.7 \\
    & IPO & 32.97 & 43.27 & 46.31 & 30.95 & 38.58 & 8.02 & 61.95 & 54.64 & 73.09 & 5.6 \\
    & KTO & 35.00 & 40.12 & 47.15 & 29.70 & 38.10 & 7.63 & 63.57 & 58.15 & 73.40 & 5.4 \\
    & CPO & 34.56 & 44.08 & 48.51 & 30.08 & 38.81 & 7.75 & 62.29 & 54.01 & 73.72 & 4.6 \\
    & SimPO & 35.09 & 43.05 & 48.95 & 31.29 & \textbf{39.15} & 8.16 & 62.80 & \textbf{60.70} & 73.32 & 3.1 \\
    & AMaPO & 35.81 & 43.69 & \textbf{49.93} & \textbf{31.47} & 38.82 & 7.82 & \textbf{65.27} & 59.23 & \textbf{76.24} & \textbf{2.2} \\
    \bottomrule
  \end{tabular}
  \vspace{-10pt}
\end{table*}

For method-specific hyperparameters, we also followed the search strategy from SimPO, noting that our method does not require any additional hyperparameters beyond the learning rate. Specifically, for different baseline methods, each with its own set of additional hyperparameters, the search strategy is in Table \ref{hyperparamater_search}. Each method is individually searched for the learning rates in [4e-7, 6e-7, 8e-7, 1e-6].  

\setlength{\tabcolsep}{2mm}
\begin{table}[h]
  \caption{The hyperparameter value in AMaPo for each training setting}
  \label{hyperparamater_AMaPO}
  \centering 
\renewcommand{\arraystretch}{0.9}
  \begin{tabular}{lcc}
    \toprule
    \textbf{Setting} & $\beta$ & Learning rate\\
    \midrule
    Llama3-8B-Base & 3.0 & 6e-7 \\
    Llama3-8B-Instruct & 3.0 & 1e-6 \\
    Mistral-7B-Base & 2.0 & 4e-7 \\
    Mistral-7B-Instruct & 2.0 & 6e-7 \\
    \bottomrule
  \end{tabular}
  \normalsize
\end{table}

\textbf{Hyperparameter in AMaPO.} 
Table \ref{hyperparamater_AMaPO} outlines the hyperparameters used for AMaPo under various settings across experiments. Our method does not introduce any extra hyperparameter expect $\beta$ inherited from the implicit reward function formulization. We have found through experience that $\beta$ can be reliably set to around 2 by default. However, the learning rate typically requires more careful tuning, which can significantly influence the performance. We recommend a slightly larger learning rate setting than SimPO, where AMaPO only assigns an adaptive margin on a subset of samples, leading to a smaller but stable gradient norm.

\textbf{Evaluation Hyperparameters. }
The hyperparameters for evaluation on AlpacaEval 2 adhere to SimPO, with a temperature of 0.7 for the Mistral-Base setting, a temperature of 0.5 for the Mistral-Instruct setting, and a temperature of 0.9 for both Llama3 settings. The hyperparameters for evaluation on MT-Bench are set according to the official decoding configuration, which specifies distinct sampling temperatures for various categories. We report the average MT-Bench score in 2 turns, and take the GPT-4-Preview-1106 as the judge model, which produces more accurate reference answers and judgments compared to GPT-4, with even lower cost \cite{meng2024simpo}.

\textbf{Computation enviroment. }
\label{sec:computation_enviroment}
All the training experiments in this paper were conducted on 4$\times$ NVIDIA A100 (80G) GPUs based on the alignment-handbook repo.

\subsection{Preference Ranking across diverse domains} \label{sec:RM-bench domain}
To thoroughly investigate the preference ranking gain of AMaPO, we report the RM-bench results across diverse domains, including Chat, Math, Code and Safety. As shown in Table~\ref{RM-bench-domain}, AMaPO consistently outperforms other baselines in various domains, especially in Chat and Safety domain. These significant improvements demonstrate that AMaPO is capable to generate more helpful and readable responses, which algins with our observations in AlpacaEval2 results in Table~\ref{main-table} and case study in Appendix~\ref{sec:case_study}.

\setlength{\tabcolsep}{2mm}
\begin{table}[h]
    \centering
    \small
    \caption{RM-bench \cite{liu2024rm} results under the four setups, including Chat, Math, Code and Safety domains. Our AMaPO can achieve good performance on distinguishing responses across various setups.}
  \label{RM-bench-domain}
\renewcommand{\arraystretch}{0.8}
  \begin{tabular}{llccccc}
    \toprule[1pt]
    & \textbf{Method} & \textbf{Avg.} & Chat & Math & Code & Safety\\
    \midrule[0.5pt]
    \multirow{5}{*}{\makecell[c]{Mistral-7B \\ Base}}
    & DPO & 55.8 & 62.7 & \textbf{55.2} & \textbf{53.0} & 52.1 \\
    & SimPO & 56.5 & 65.6 & 54.7 & 53.0 & 52.5 \\
    & $\alpha$-DPO & \textbf{58.4} & 66.9 & 54.4 & 52.0 & \textbf{60.3} \\
    & AMaPO & 58.1 & \textbf{67.5} & 55.0 & 52.5 & 57.4\\
     \midrule[0.8pt]
    \multirow{5}{*}{\makecell[c]{Llama3-8B \\ Base}}
    & DPO & 54.6 & 60.6 & 54.8 & \textbf{53.3} & 54.6\\
    & SimPO & 56.9 & 67.6 & 55.2 & 52.3 & 52.7 \\
    & $\alpha$-DPO & 58.2 & 69.4 & 55.2 & 52.8 & 55.6 \\
    & AMaPO & \textbf{58.6} & \textbf{69.5} & \textbf{55.5} & 53.2 & \textbf{56.2} \\
     \midrule[0.8pt]
    \multirow{5}{*}{\makecell[c]{Mistral-7B \\ Instruct}}
    & DPO & 53.2 & 56.1 & 53.6 & \textbf{52.4} & 50.8\\
    & SimPO & 54.9 & \textbf{61.2} & \textbf{54.4} & \textbf{52.4} & 51.5 \\
    & $\alpha$-DPO & 55.2 & 59.3 & 53.7 & 51.9 & \textbf{55.9} \\
    & AMaPO & \textbf{55.5} & 60.7 & 54.2 & 52.0 & 55.0 \\
    \midrule
    \multirow{5}{*}{\makecell[c]{Llama3-8B \\ Instruct}}
    & DPO & 53.4 & 60.5 & 54.1 & 51.2 & 47.6 \\
    & SimPO & 55.7 & 64.9 & 55.0 & 52.7 & 50.3 \\
    & $\alpha$-DPO & 55.0 & 62.5 & \textbf{55.5} & 52.3 & 49.4 \\
    & AMaPO & \textbf{56.5} & \textbf{65.9} & 55.0 & \textbf{53.6} & \textbf{51.7} \\
    \bottomrule[1pt]
  \end{tabular}
  \vspace{-10pt}
\end{table}
\subsection{Downstream Task Evaluation} \label{sec:Downstream_tasks}
To further examine the performance of the policy model on downstream tasks across baselines, we also evaluate models fine-tuned with different methods on the same dataset across popular tasks listed on the Huggingface Open LLM Leaderboard v1 and v2\footnote{\url{https://huggingface.co/open-llm-leaderboard}} following previous works \cite{xiao2024cal, xiao2025simper}. Open LLM Leaderboard encompasses ten tasks, including reasoning benchmarks (MMLU-PRO\cite{wang2024mmlu}, Winogrande\cite{sakaguchi2021winogrande}, ARC\cite{clark2018think}, IFEval\cite{zhou2023instruction}, BBH\cite{suzgun2022challenging}, GPQA\cite{rein2024gpqa}), mathematical reasoning (MATH\cite{hendrycks2021measuring}), and safety benchmark (TruthfulQA\cite{lin2021truthfulqa}). We follow the established evaluation protocols provided by the Language Model Evaluation Harness library\cite{eval-harness} and present the results in Table \ref{downstream-table}.

\textbf{Gain Significant improvements in Avg. rank. }
As shown in Avg. Rank column, AMaPo achieves the best performance across various downstream tasks and setups mostly. Thess consistent and significant improvements highlight the robustness and effectiveness of our method, which suggest that enhanced preference ranking capabilities can help the policy model complete downstream tasks.

\textbf{Reading comprehension and commonsense reasoning improve.} For ARC and Winograd, which require strong comprehension and reasoning capabilities, AMaPO achieves remarkable improvements over baselines. We hypothesis that our method helps the policy model to capture and understand the subtle differences in responses, aligning with the observations on RM-Bench (see Table \ref{RM-bench-table}), then generalize to better reasoning performance.

\textbf{Math performance drops. }
However, although a significant improvement of AMaPO is observed on the ranking accuracy in code and math domain, AMaPO still exhibits a huge drop in MATH. We hypothesize that there is a gap between identifying and generating an accurate response. Specifically, AMaPO consistently underperforms SimPO in MATH, which suggests that assigning a positive margin for correctly ranked samples may help the generation performance on math tasks. These findings emphasize the importance of the margin design, and further exploration is left to future work.

\setlength{\tabcolsep}{4pt}
\begin{table}[h]
    \centering
    \small
    \caption{Ablation study of oracle margin estimation under Llama3-8B-Base setup.}
  \label{bsz_ablation}
  \renewcommand{\arraystretch}{1.0}
    \begin{tabular}{lcccc}
    \toprule
    \multirow{2}{*}{\makecell{\textbf{Batch} \\ \textbf{Size}}} & \multicolumn{2}{c}{\textbf{RM-Bench}} & \multicolumn{2}{c}{\textbf{AlpacaEval2}} \\
    \cmidrule(r){2-3} \cmidrule(r){4-5}
    & \textbf{\footnotesize Easy} \footnotesize(\%) & \textbf{\footnotesize Hard} \footnotesize (\%) & \textbf{\footnotesize LC} \footnotesize(\%) & \textbf{\footnotesize WR} \footnotesize (\%) \\
    \midrule
    16 & \textbf{88.2} & 24.5 & \textbf{27.9} & \textbf{23.5} \\
    32 (default) & 87.8 & 25.4 & 26.4 & 21.4 \\
    64 & 87.5 & \textbf{25.8} & 22.1 & 20.8 \\
    \bottomrule[1pt]
    \end{tabular}
  \vspace{-10pt}
\end{table}


\subsection{Ablation Study of Oracle Margin Estimation}
\label{app: Ablation Study of Oracle Margin Estimation}
To validate the robustness of our proposed estimation methods based on the mean implicit reward margin within the batch, we conduct ablation study towards the batch size hyperparameter. To ensure a fair comparison, we adjusted the batch size for oracle margin estimation while holding the total number of gradient update steps constant (the same as default setup) for under Llama3-8B-Base setup on Ultrafeedback Binarized dataset. As shown in Table~\ref{bsz_ablation}, results on ranking accuracy and downstream generation benchmarks, demonstrate the robustness of our oracle margin estimation. Specifically, an increasing batch size leads to a performance drop on the RM-Bench Easy accuracy but a gain on the Hard accuracy. This implies that AMaPO amplifies the gradients for hard, incorrectly ranked samples, demonstrating that a larger batch size facilitates a more precise estimation of the oracle margin. However, results on AlpacaEval2 show an opposite trend. This is consistent with our ablation studies over $\beta$ in Section~\ref{sec:analysis}, which indicates that excessively amplifying the gradients of hard samples leads to distribution sharpening, resulting in a degradation of downstream performance. Our analysis reveals that AMaPO not only consistently outperforms strong baselines (Table~\ref{main-table}) irrespective of the batch size, but its performance sensitivity across diverse datasets also provides compelling empirical evidence for the two limiting cases predicted by our theory, thereby validating both the practical effectiveness and theoretical soundness of our approach.

\begin{figure}[h]
\centering
    \centering
        \includegraphics[width=0.32\textwidth]{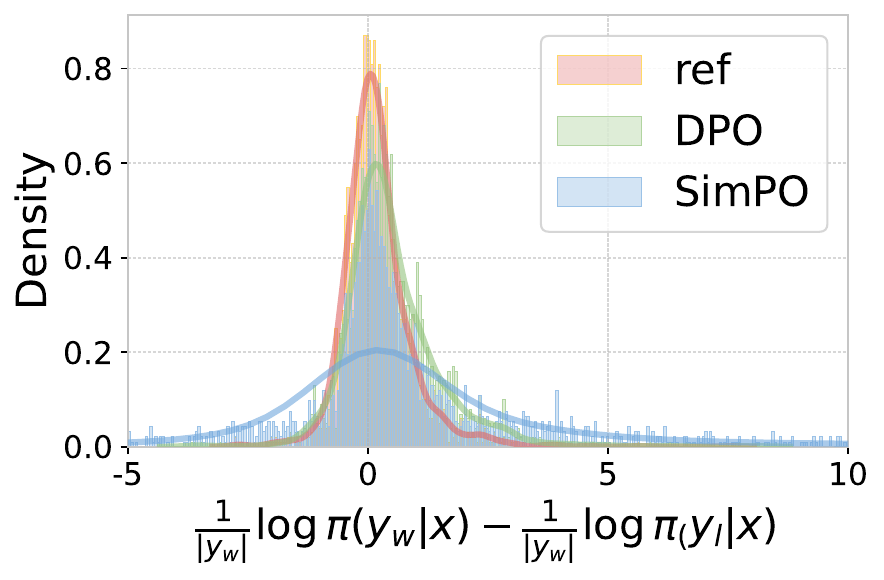}
        \caption{Reward margin on the test set of Ultrafeedback Binarized.
        \label{fig:reward_margin}}
\end{figure}
\vspace{-5mm}

\subsection{Reward Distribution of Different Models}
\label{app: Reward Distribution of Different Models}
To validate that a tuned policy model is a better option to estimate the oracle reward margin, we visualize the distribution of reward margin on Ultrafeedback Binarized dataset under Llama3-Base-8B setup with popular methods (i.e. DPO and SimPO). As shown in Figure \ref{fig:reward_margin}, tuned policy models, SimPO specifically, exhibit a rightward skew distribution with a larger variance, suggesting that tuned policy models are more likely to produce accurate oracle reward margin.

\setlength{\tabcolsep}{2pt} 
\begin{table*}[!h]
    \vspace{-5pt}
    \centering
    \small
    \caption{Detailed results of ablation studies on AlpacaEval2.}
  \label{length_variants}
\renewcommand{\arraystretch}{0.9}
\begin{tabular}{lcccccccc}
    \toprule
    \multirow{2}{*}{\textbf{Method}} & \multicolumn{4}{c}{\textbf{Llama3-8B-Base}} & \multicolumn{4}{c}{\textbf{Llama3-8B-Instruct}} \\
    \cmidrule(r){2-5} \cmidrule(r){6-9}
    & \textbf{\footnotesize LC (\%)} & \textbf{\footnotesize WR} \footnotesize (\%) & \textbf{\small STD (\%)} & \textbf{\small Length} & \textbf{\footnotesize LC} \footnotesize(\%) & \textbf{\footnotesize WR} \footnotesize (\%) & \textbf{\small STD (\%)} & \textbf{\small Length} \\
    \midrule
    AMaPO & 26.4 & 21.4 & 1.3 & 1592 & 46.1 & 41.3 & 1.5 & 1776\\
    \midrule 
    w/o Z-norm & 24.8 & 20.4 & 1.4 & 1754 & 44.2 & 39.6 & 1.6 & 1754\\ 
    w/o exp & 24.0 & 17.6 & 1.2 & 1480 & 47.1 & 42.9 & 1.6 & 1783\\ 
    w/o adaptive & 20.7 & 16.4 & 1.1 & 1493 & 47.2 & 42.5 & 1.6 & 1790\\ 
    w/o zero & 22.3 & 21.1 & 1.3 & 1918 & 48.6 & 45.0 & 1.6 & 1828\\
    \bottomrule
    \end{tabular}
    \vspace{-5pt}
\end{table*}
\begin{table*}[!h]
  \caption{Detailed results of AlpacaEval2. LC means length-controlled win tate, WR means raw win rate, and STD means standard deviation of win rate. Length is the average generation length.}
  \label{detailed-alpaca}
  \centering 
\renewcommand{\arraystretch}{0.9}
  \begin{tabular}{lcccccccc}
    \toprule
    \multirow{2}{*}{\textbf{Method}} & \multicolumn{4}{c}{\textbf{Llama3-8B-Base}} & \multicolumn{4}{c}{\textbf{Llama3-8B-Instruct}} \\
    \cmidrule(r){2-5} \cmidrule(r){6-9}
    & \textbf{\footnotesize LC} \footnotesize(\%) & \textbf{\footnotesize WR} \footnotesize (\%) & \textbf{\small STD (\%)} & \textbf{\small Length} & \textbf{\footnotesize LC} \footnotesize(\%) & \textbf{\footnotesize WR} \footnotesize (\%) & \textbf{\small STD (\%)} & \textbf{\small Length}\\
    \midrule
    SFT & 6.2 & 4.6 & 0.7 & 1082 & 26.0 & 25.3 & 1.3 & 1920\\
    DPO & 18.2 & 15.5 & 1.1 & 1585 & 40.3 & 37.9 & 1.4 & 1883\\
    SLiC & 12.3 & 13.7 & 0.9 & 1540 & 26.9 & 27.5 & 1.3 & 1977\\
    IPO & 14.4 & 14.2 & 1.1 & 1856 & 35.6 & 35.6 & 1.4 & 1983\\
    KTO & 14.2 & 12.4 & 1.0 & 1646 & 33.1 & 31.8 & 1.4 & 1909\\
    CPO & 10.8 & 8.1 & 1.0 & 2495 & 28.9 & 32.2 & 1.4 & 2166\\
    SimPO & 22.0 & 20.3 & 1.2 & 1795 & 44.7 & 40.5 & 1.4 & 1825\\
    $\alpha$-DPO & 21.7 & 20.6 & 1.3 & 1913 & 46.6 & 39.6 & 1.6 & 1759\\
    AMaPO & 26.4 & 21.4 & 1.3 & 1592 & 46.1 & 41.3 & 1.5 & 1776\\
    \toprule
    \multirow{2}{*}{\textbf{Method}} & \multicolumn{4}{c}{\textbf{Mistral-7B-Base}} & \multicolumn{4}{c}{\textbf{Mistral-7B-Instruct}} \\
    \cmidrule(r){2-5} \cmidrule(r){6-9}
    & \textbf{\footnotesize LC} \footnotesize(\%) & \textbf{\footnotesize WR} \footnotesize (\%) & \textbf{\small STD (\%)} & \textbf{\small Length} & \textbf{\footnotesize LC} \footnotesize(\%) & \textbf{\footnotesize WR} \footnotesize (\%) & \textbf{\small STD (\%)} & \textbf{\small Length}\\
    \midrule
    SFT & 8.4 & 6.2 & 1.1 & 914 & 17.1 & 14.7 & 1.1 & 1676 \\
    DPO & 15.1 & 12.5 & 1.0 & 1477 & 26.8 & 24.9 & 1.3 & 1808\\
    SLiC & 10.9 & 8.9 & 0.9 & 1525 & 24.1 & 24.6 & 1.3 & 2088\\
    IPO & 11.8 & 9.4 & 0.9 & 1380 & 20.3 & 20.3 & 1.2 & 2024\\
    KTO & 13.1 & 9.1 & 0.9 & 1144 & 24.5 & 23.6 & 1.3 & 1901\\
    CPO & 9.8 & 8.9 & 0.9 & 1827 & 23.8 & 28.8 & 1.3 & 3245\\
    SimPO & 21.5 & 20.8 & 1.2 & 1868 & 32.1 & 34.8 & 1.4 & 2193\\
    $\alpha$-DPO & 17.2 & 13.0 & 1.1 & 1200 & 34.2 & 33.8 & 1.6 & 1980\\
    AMaPO & 24.3 & 20.6 & 1.3 & 1700 & 34.5 & 35.1 & 1.5 & 2121\\
    \bottomrule
  \end{tabular}
  \vspace{-10pt}
\end{table*}

\subsection{Generation Length Analysis} \label{sec:Length_Analysis}
\textbf{AMaPO decreases generation length and improves generation quality.} As we have observed that AMaPO exhibits a significant on AlpacaEval2 LC win rate improvement compared to other baselines, to further analyze the underlying working inners, we first report the detailed results of AlpacaEval2 including the strandard deviation and generation length. As shown in Table \ref{detailed-alpaca}, although AMaPO outperforms baselines on WR slightly, it generates response extremely shorter than comparable baselines. Notably, AMaPO decreases the generation length by up to 203 tokens compared to SimPO. This suggests that AMaPO can effectively control the verbosity of the generated responses, and meanwhile improve the generation quality to a certain extent. Also, we find that the designs of adaptive margin profoundly impact the generation length, as shown in Table \ref{length_variants}. Specifically, removing the zero setting for correctly ranked samples increases the generation length significantly. We argue that it is caused by learning the preference patterns from the already correctly ranked samples, in which preferred responses tend to be longer in existing preference datasets.

\section{Case Study} \label{sec:case_study}
In this section, we present examples where Llama-Instruct-AMaPO generates a better-structured answers compared to Llama-Instruct-SimPO in AlpacaEval2 benchmark.

Given the question, ``What are the names of some famous actors that started their careers on Broadway?", the SimPO model generates a response that only contains a list of celebrity names, lacking corresponding explanatory material. In contrast, the AMaPO model produces a detailed answer that not only lists celebrity names but also covers their famous acting experiences. This helps users learn about celebrities faster and makes it easier to verify the information's authenticity, indicating that AMaPO can better leverage its internal knowledge.

Similarly, given the question, ``How to improve instance segmentation AP metrics with human knowledge", the SimPO model generates a response with a long list of bullet points, making it difficult to understand the relationships between different points. In contrast, the AMaPO model produces a well-structured answer with high-level categorization of different behaviors, organized with a General-Specific-General structure with the markdown form. This makes the answer more readable and easier to understand.

\begin{table*}[]
    \centering
        \begin{tabular}{lp{11.5cm}}
        \toprule[1pt]
        \bf{Prompt} & \texttt{What are the names of some famous actors that started their careers on Broadway?}\\
        \midrule
        \bf{AMaPO} & 
        Many talented actors have transitioned from Broadway to Hollywood success. Here are some famous examples:\newline 1. Julie Andrews: Starred in Broadway's "My Fair Lady" (1956) and Camelot (1960) before becoming a movie icon in films like The Sound of Music (1965) and Mary Poppins (1964).\newline2. Hugh Jackman: Originated the role of Curly in Broadway's Oklahoma! (1998) and later starred in films like X-Men (2000), Les Misérables (2012), and The Greatest Showman (2017).\newline3. Neil Patrick Harris: Got his start on Broadway in Assassins (1990) and Rent (1996) before becoming a TV star in Doogie Howser, M.D. and later How I Met Your Mother.\newline4. Idina Menzel: Originated the role of Maureen Johnson in Broadway's Rent (1996) and later voiced Elsa in Disney's Frozen (2013) and starred in TV's Glee.\newline5. Lin-Manuel Miranda: Created and starred in Broadway's In the Heights (2008) and Hamilton (2015), earning multiple Tony Awards and a film career in Moana (2016) and Mary Poppins Returns (2018).\newline6. Nathan Lane: Began his career on Broadway in Noises Off (1983) and Guys and Dolls (1992), later appearing in films like The Birdcage (1996) and TV's Modern Family.\newline7. Bernadette Peters: Starred in Broadway's Song and Dance (1985) and Into the Woods (1987), with notable film roles in The Muppet Movie (1979) and TV's Smash.\newline8. Alan Cumming: Originated the role of Emcee in Broadway's Cabaret (1998) and has since appeared in films like X2: X-Men United (2003), Spy Kids (2001), and TV's The Good Wife.\newline9. Sutton Foster: Got her start on Broadway in Annie (1997) and Thoroughly Modern Millie (2002), later starring in TV's Bunheads and Younger.\newline10. Laura Linney: Began her career on Broadway in Six Degrees of Separation (1990) and The Little Foxes (1997), with notable film roles in The Truman Show (1998) and TV's Ozark.\newline These are just a few examples of talented actors who started their careers on Broadway and went on to achieve success in film and television.
        \\
        \midrule
        \bf{SimPO} & 
    1. Hugh Jackman \newline 2. Audra McDonald \newline 3. Bernadette Peters\newline4. Angela Lansbury\newline5. Meryl Streep\newline6. Matthew Broderick\newline7. Nathan Lane\newline8. Patti LuPone\newline9. Lin-Manuel Miranda\newline10. Glenn Close\newline11. Cynthia Nixon\newline12. Liev Schreiber\newline13. Sutton Foster\newline14. James Earl Jones\newline15. Phylicia Rashad\newline16. Leslie Odom Jr.\newline17. Jessie Mueller\newline18. Kelli O'Hara\newline19. Brian Stokes Mitchell\newline20. Bernadette Peters (again, as mentioned earlier)",
        \\

        \bottomrule[1pt]
    \end{tabular}
    \caption{An AlpacaEval 2 generation from the Llama-3-8B-Instruct model trained with \texttt{AMaPO} and \texttt{SimPO}. Weighted-Alpaca-Eval-GPT4-Turbo prefers \texttt{AMaPO} over \texttt{SimPO}. Compared to \texttt{SimPO}, the generation by \texttt{AMaPO} contains more related information, with hierarchical discussions that make it more helpful. 
    }
    \label{tab:case-1}
\end{table*}

\begin{table*}[]
    \centering
        \begin{tabular}{lp{11.5cm}}
        \toprule[1pt]
        \bf{Prompt} & \texttt{How to improve instance segmentation AP metrics with human knowledge}\\
        \midrule
        \bf{AMaPO} & 
            Instance segmentation models, such as Mask R-CNN, typically evaluate their performance using metrics like Average Precision (AP) and Intersection over Union (IoU). While these metrics provide a quantitative measure of performance, incorporating human knowledge can help improve instance segmentation models in several ways:\newline
            1. \textbf{Annotate challenging cases}: Identify and annotate instances with complex or ambiguous boundaries, occlusions, or unusual appearances. This helps the model learn to handle these edge cases better.\newline
            2. \textbf{Provide domain-specific knowledge}: Incorporate domain expertise into the annotation process by highlighting important features, such as texture, color, or shape, that are relevant to the specific application (e.g., medical imaging, autonomous driving).\newline
            3. \textbf{Use weak supervision}: Instead of annotating every instance, use weak supervision techniques like: 
            \begin{itemize}[label=-]
                \item Bounding box annotations: Provide approximate bounding boxes around instances, allowing the model to learn from rough estimates.
                \item Semantic segmentation masks: Annotate classes or categories, rather than individual instances, to help the model understand the context and relationships between objects.
            \end{itemize}
            4. \textbf{Regularly update and refine annotations}: As the model improves, revisit and refine the annotations to ensure they remain accurate and relevant. This helps the model adapt to its own mistakes and learn from corrections.\newline
            5. \textbf{Use transfer learning and multi-task learning}: Initialize the model with pre-trained weights on a related task or dataset, and then fine-tune it on your specific instance segmentation task. This can leverage knowledge from other domains or tasks to improve performance.\newline
            6. \textbf{Incorporate prior knowledge into the model architecture}: Design the model to incorporate domain-specific constraints or features, such as: 
            \begin{itemize}[label=-]
                \item Using attention mechanisms to focus on relevant regions or features.
                \item Incorporating shape priors or geometric constraints to improve instance segmentation.
            \end{itemize} 
            7. \textbf{Use human-in-the-loop evaluation}: Involve human evaluators in the evaluation process to: 
            \begin{itemize}[label=-]
                \item Provide feedback on model performance and identify areas for improvement. 
                \item Correct mistakes and annotate challenging cases.
                \item Validate the model's output and ensure it aligns with human expectations.
            \end{itemize}
            \textbf{Explore alternative evaluation metrics}: Consider using metrics that better capture the nuances of instance segmentation, such as: 
            \begin{itemize}[label=-]
                \item Instance-level metrics like mean Average Precision (mAP) or mean Average Recall (mAR).
                \item Metrics that account for instance size, orientation, or aspect ratio.
            \end{itemize}
            9. \textbf{Regularly monitor and analyze model behavior}: Use visualization tools and techniques to understand how the model is making predictions, identifying biases, and areas where it struggles. This helps you to refine the model and annotations accordingly. \newline
            10. \textbf{Collaborate with domain experts}: Work closely with experts in the domain to ensure the model is aligned with their understanding of the problem and to incorporate their insights into the annotation process and model development.\newline
            By incorporating these strategies, you can leverage human knowledge to improve the performance and robustness of your instance segmentation model, ultimately leading to better AP metrics and more accurate results.

        \\
        \midrule
    \end{tabular}
    \end{table*}
    
    \begin{table*}[!h]
        \centering
        \begin{tabular}{lp{11.5cm}}
            \midrule
        \bf{SimPO} & 
            1. \textbf{Incorporate Prior Knowledge}: Incorporating domain-specific knowledge, such as object categories, shapes, textures, and common configurations, can enhance instance segmentation models' performance. For instance, leveraging pre-existing object detection models or semantic segmentation models can provide additional context and guidance for instance segmentation.\newline
            2. \textbf{Collaborative Learning}: Collaborative learning involves incorporating human feedback into the training process. This approach involves collecting annotations from human experts, such as bounding boxes or segmentation masks, for a subset of images, and using them to guide the model's learning process. This feedback can help the model better understand object boundaries, relationships, and nuances, leading to improved segmentation accuracy.\newline
            3. \textbf{Active Learning}: Active learning involves selecting the most informative samples for annotation based on the model's uncertainty or confidence scores. By prioritizing challenging instances, human annotators can focus their efforts on areas where the model struggles, leading to more targeted feedback and improved instance segmentation performance.\newline
            4. \textbf{Multi-Task Learning}: Combining instance segmentation with other relevant tasks, such as object detection, semantic segmentation, or panoptic segmentation, can leverage shared features and enhance the model's overall understanding of the scene. This approach can improve the segmentation accuracy by providing additional supervision signals and reducing the semantic gap between different tasks.\newline
            5. \textbf{Domain Adaptation}: Instance segmentation models trained on synthetic datasets or generic benchmarks may not generalize well to real-world scenarios with varying backgrounds, lighting conditions, and object appearances. Incorporating domain-specific knowledge, such as transfer learning from pre-trained models, data augmentation techniques, or adversarial training, can help adapt the model to specific domains, leading to better instance segmentation performance.\newline
            6. \textbf{Ensemble Methods}: Combining multiple instance segmentation models with diverse architectures, training strategies, or loss functions can leverage their strengths and mitigate their weaknesses, resulting in improved overall performance. Ensemble methods, such as voting or weighted averaging, can aggregate predictions from multiple models, reducing uncertainty and increasing robustness.\newline
            7. \textbf{Post-Processing Techniques}: Applying post-processing techniques, such as non-maximum suppression, merging overlapping segments, or refining segmentation masks based on contextual cues, can enhance the segmentation quality and reduce false positives\/negatives. Human knowledge can guide the selection of appropriate thresholds, heuristics, or priors for these techniques, leading to more accurate instance segmentation results.\newline
            8. \textbf{Interactive Feedback Loops}: Interactive feedback loops involve iteratively refining the segmentation results with human feedback, allowing for continuous improvement and adaptation. This approach involves presenting segmentation outputs to human annotators, soliciting feedback, and incorporating it back into the model's training pipeline. This iterative process can lead to more accurate segmentation results tailored to specific use cases or applications.
        \\
        \bottomrule[1pt]
    \end{tabular}
    \caption{An AlpacaEval 2 generation from the Llama-3-8B-Instruct model trained with \texttt{AMaPO} and \texttt{SimPO}. Weighted-Alpaca-Eval-GPT4-Turbo prefers \texttt{AMaPO} over \texttt{SimPO}. Compared to \texttt{SimPO}, the generation by \texttt{AMaPO} is organized with a General-Specific-General structure, with markdown form makes it easier to read. 
    }
    \label{tab:case-1}
\end{table*}

\end{document}